%% file: main.tex
\def\year{2024}\relax
\documentclass[letterpaper]{article} 
\usepackage{aaai24}  
\usepackage{times}  
\usepackage{helvet}  
\usepackage{tikz}
\usepackage{courier}  
\usepackage[hyphens]{url}  
\usepackage{graphicx} 
\usepackage{graphicx}
\usepackage{subfigure}
\usepackage{booktabs} 
\usepackage{centernot}
\usepackage{enumitem}
\usepackage{multicol}
\usepackage[utf8]{inputenc} 
\usepackage[T1]{fontenc}    
\usepackage{booktabs}       
\usepackage{amsfonts}       
\usepackage{nicefrac}       
\usepackage{xcolor}         
\usepackage{tikz}
\usepackage{url}
\usepackage{verbatim}
\usepackage{mathtools}
\usepackage{multirow}
\usepackage{amsthm}
\usepackage{amssymb}
\usepackage{pifont}
\newcommand{\cmark}{\ding{51}}%
\newcommand{\xmark}{\ding{55}}%
\usepackage{thmtools}
\usepackage{thm-restate}
\usepackage{caption}
\usepackage{enumitem}
\usepackage{mdframed}
\newmdenv[
  topline=false,
  bottomline=false,
  rightline=false,
  linewidth=0.3mm,
  skipabove=1mm,
  skipbelow=1mm
]{siderules}

\usepackage{booktabs}
 
\usepackage{amssymb}
\usepackage{float}
\usepackage{amsmath}
\usepackage{xspace}
\definecolor{r}{rgb}{0, 0, 0}

\newtheorem{proposition-informal}{Proposition (Informal)}[section]
\newtheorem{definition}{Definition}[section]
\newtheorem{assumption}{Assumption}[section]
\newtheorem{hypothesis}{Hypothesis}[section]
\usepackage[numbers]{natbib}  
\usepackage{caption} 
\usepackage{color, colortbl}
\defcitealias{sundararajan2017axiomatic}{S 2017}
\defcitealias{ca}{AC 2019}
\defcitealias{credo}{SK 2022}
\defcitealias{causalshapley}{TH 2020}

\definecolor{Gray}{gray}{0.9}

\DeclareCaptionStyle{ruled}{labelfont=normalfont,labelsep=colon,strut=off} 
\usepackage{algorithm}
\usepackage{algorithmic}

\usepackage{newfloat}
\usepackage{listings}
\floatstyle{ruled}
\newfloat{listing}{tb}{lst}{}
\floatname{listing}{Listing}

\title{Towards Learning and Explaining Indirect Causal Effects in Neural Networks}
\author {
    Abbavaram Gowtham Reddy\textsuperscript{\rm 1},
    Saketh Bachu\textsuperscript{\rm 1},
    Harsharaj Pathak\textsuperscript{\rm 1},
    Benin L. Godfrey\textsuperscript{\rm 1},
     Varshaneya V\textsuperscript{\rm 2},
    Vineeth N. Balasubramanian\textsuperscript{\rm 1},
    Satyanarayan Kar\textsuperscript{\rm 2}
}
\affiliations {
    \textsuperscript{\rm 1} Indian Institute of Technology Hyderabad, India\\
    \textsuperscript{\rm 2} Honeywell, Bengaluru, India\\
}

\begin{document}

\maketitle

\begin{abstract}
Recently, there has been a growing interest in learning and explaining causal effects within Neural Network (NN) models. By virtue of NN architectures, previous approaches consider only direct and total causal effects assuming independence among input variables. We view an NN as a structural causal model (SCM) and extend our focus to include indirect causal effects by introducing feedforward connections among input neurons. We propose an ante-hoc method that captures and maintains direct, indirect, and total causal effects during NN model training. We also propose an algorithm for quantifying learned causal effects in an NN model and efficient approximation strategies for quantifying causal effects in high-dimensional data. Extensive experiments conducted on synthetic and real-world datasets demonstrate that the causal effects learned by our ante-hoc method better approximate the ground truth effects compared to existing methods.
\end{abstract}

\section{Introduction}
\label{sec introduction}

Neural network (NN) models enriched with causal knowledge have demonstrated their ability to achieve robustness~\citep{scholkopf2021toward}, invariance~\citep{icm,rim}, and provide interpretable explanations for human understanding~\citep{ca,causalexplanationblackbox,credo}. In training such NN models imbued with causal knowledge, two primary tasks emerge: (1) acquiring a comprehension of causal relationships between input and output neurons~\citep{janzing2019causal,kyono2020castle,credo}, and (2) validating and explaining the acquired causal relationships~\citep{ca,acausalproblem,causalexplanationblackbox}. Previous studies have tended to address these two tasks separately, despite their close interconnectedness. This separation of dependent tasks also makes it challenging to study and model more nuanced aspects such as the indirect causal effects of input neurons on the output of an NN. To address this limitation, in this work, we propose an \underline{A}nte-\underline{H}oc \underline{C}ausal \underline{E}xplanations (AHCE) approach that simultaneously performs both these tasks.

\begin{figure}
    \centering
    \scalebox{0.45}{
    \tikzset{every picture/.style={line width=0.75pt}}  
    \input{images/introduction.tikz}
    }
    \caption{\footnotesize (a) A marginalized NN whose inputs $S, E, R$ are not causally related. (b) A marginalized NN whose inputs are connected through feedforward connections (e.g., $S \rightarrow E$) to capture underlying causal relationships (e.g., $S$ causes $E$) to learn the indirect causal effects of inputs on output (e.g. effect of $S$ on $I$ via $E$).}
    \label{fig:introduction}
\end{figure}
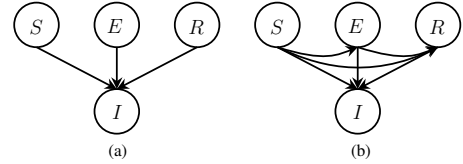
\noindent \textbf{Task 1 - Learning Causal Effects in NNs:} A common practice in learning causal effects in NN models involves considering the NN as a Structural Causal Model (SCM), representing the parametric causal relationships between the features~\citep{kocaoglu2017causalgan,ca,acausalproblem}. Given our focus on input-output causal relationships in an NN, following~\citep{kocaoglu2017causalgan,ca,credo}, we marginalize the hidden layers and view the output as a function of inputs as shown in  Fig~\ref{fig:introduction} (a) (the motivating example in the next paragraph describes the variables). It becomes evident that the SCM embodied by a conventional feedforward NN model lacks causal relationships among input features (neurons in the first layer, we use \textit{input features} and \textit{input neurons} interchangeably in this work). Consequently, the causal effects that are learned and quantified are restricted solely to \textit{direct causal effects} (\textit{viz.} causal effects that do not propagate through other input features -- see Appendix \S\ref{sec preliminaries} for preliminaries). Hence, there is currently no feasible approach for explaining \textit{indirect causal effects} (\textit{viz.} causal effects that propagate through other input features). We extend the basic architecture of an NN by adding feedforward connections among input neurons (see Fig~\ref{fig:introduction}(b)) based on domain knowledge of how features interact in the real-world, thus enabling the learning and explaining of indirect causal effects.

To motivate the need for the study of indirect causal effects in NN models, consider the task of predicting an individual's income ($I$) using the features: education ($E$), socioeconomic status ($S$), and job role ($R$). In the real world, $S$ causes $E$ and $R$; $E$ causes $R$; $S, E,$ and $R$ cause $I$ (see Fig~\ref{fig:introduction}(b)). However, in an NN model, the relationships among input features $S, E, R$ are not modeled (see Fig~\ref{fig:introduction}(a)). As a result, for feature $S$, an NN model can only learn and explain direct causal effects while neglecting the indirect causal effects on $I$ propagating via $E$ and $R$. From a fairness standpoint, $S$ should have no direct causal effect on $I$ but can exhibit a non-zero indirect causal effect on $I$ through $E$ and $R$. 
If a model learns a non-zero direct causal effect of $S$ on $I$, the corresponding model explanations may not align with the real-world and can indicate unacceptable learned causal effects. Thus, learning indirect effects can also find application in identifying and comprehending model biases. We provide the ability to differentiate between direct and indirect causal effects in an NN model by introducing feedforward connections among input features (see Appendix \S\ref{sec another motivating example} for another motivating example).

\begin{table}
    \centering 
    \scalebox{0.73}{
    \begin{tabular}{lccccc}
    \toprule
         & \textbf{Direct} & \textbf{Indirect} &\textbf{Total}&\textbf{Causal}&\textbf{Ante-hoc}\\
       \textbf{Method} &\textbf{Effects}&\textbf{Effects}&\textbf{Effects}&\textbf{Effects}&\textbf{Explanations}\\
    \midrule
        IG {\footnotesize \citepalias{sundararajan2017axiomatic}} & \textcolor{blue}{\textcolor{blue}{\Large \cmark}} &\textcolor{red}{\large \xmark} &\textcolor{red}{\Large \xmark} & \textcolor{red}{\Large \xmark} &\textcolor{red}{\Large \xmark} \\
        CA {\footnotesize \citepalias{ca}}& \textcolor{blue}{\Large \cmark} &\textcolor{red}{\Large \xmark} &\textcolor{red}{\Large \xmark} & \textcolor{blue}{\Large \cmark} &\textcolor{red}{\Large \xmark} \\
        CSHAP {\footnotesize \citepalias{causalshapley}}&\textcolor{blue}{\Large \cmark} &\textcolor{blue}{\Large \cmark} &\textcolor{blue}{\Large \cmark} &\textcolor{red}{\Large \xmark} &\textcolor{red}{\Large \xmark} \\
        CREDO {\footnotesize \citepalias{credo}} &\textcolor{blue}{\Large \cmark} &\textcolor{red}{\Large \xmark} &\textcolor{blue}{\Large \cmark} &\textcolor{blue}{\Large \cmark} &\textcolor{blue}{\Large \cmark} \\
        \midrule
        \textbf{AHCE {\footnotesize (Ours)}}&\textcolor{blue}{\Large \cmark} &\textcolor{blue}{\Large \cmark} &\textcolor{blue}{\Large \cmark} &\textcolor{blue}{\Large \cmark} &\textcolor{blue}{\Large \cmark}\\
    \bottomrule
    \end{tabular}
    }
    \caption{\footnotesize Comparison of various explanation methods. IG refers to Integrated Gradients and CA refers to Causal Attributions.} 
    \label{tab:comparison}
\end{table} 

\noindent \textbf{Task 2 - Explaining Causal Effects in NNs:} Explainability methods for NN models have encompassed a wide range of techniques ranging from various gradient-based methods to Shapley values. Recently, there has been increased attention towards causal explanations due to their enhanced reliability~\citep{wachter,Hendricks_2018}, as well as their potential for aiding in debugging~\citep{lmdebugger} and improving NN model performance~\citep{kyono2020castle, credo}. We refer to explanations such as gradients and Shapley values as \textit{effects} and causal explanations as \textit{causal effects} to separate the non-causal explanations from causal explanations. Most explanation methods provide direct effects, such as gradients and marginal Shapley values~\citep{shapley}. Causal Shapley values (CSHAP)~\citep{causalshapley} account for indirect effects mediated through other features. However, they are not equal to the causal effects obtained through backdoor adjustment~\citep{pearl2009causality} (see Appendix \S\ref{sec cshap comparison} for details). Except for causal regularization using domain priors (CREDO)~\citep{credo}, all existing efforts in causal explanations are post-hoc approaches, quantifying the causal effects of input features on the output for a pre-trained NN model. These post-hoc explanation methods, though causal, only capture direct effects, and assign zero indirect causal effects to all features. This may not accurately represent the true underlying indirect causal effects among input features in the real world~\citep{acausalproblem}. Although~\citep{credo} adopts an ante-hoc approach, it does not model indirect causal effects. See Tab~\ref{tab:comparison} for a comparison of related explanation methods. To the best of our knowledge, this is the first work that that provides an ante-hoc approach to explain indirect causal effects. Our key contributions are summarized below.
\begin{itemize}[leftmargin=*]
 \setlength\itemsep{-0.2em}
    \item We propose a novel ante-hoc training algorithm to capture indirect causal effects in NN models. Our approach aligns with the demand for intrinsically interpretable techniques rather than post-hoc explanations~\citep{rudin2021interpretableml}.
    \item We propose an algorithm to quantify the learned indirect causal effects in NNs using the lateral connections among input neurons.
    \item We also present effective implementation strategies to scale causal explanation methods to high-dimensional data w.r.t. time and space complexity.
    \item We present a wide range of empirical results on both synthetic and real-world datasets to showcase the usefulness of the proposed method.
\end{itemize}  

\section{Related Work}
\label{sec related work}

\noindent \textbf{Learning Structural Causal Models:}
Learning the structural causal model (SCM) is a core component of tasks in causal inference, including causal effect estimation~\citep{xia2021causal}, and counterfactual generation~\citep{deepscm}. In a work possibly closest to our work,~\citep{xia2021causal} propose the learning of neural causal models (NCM) utilizing the underlying causal graph as an inductive bias, with a specific emphasis on identifying and learning ground truth causal effects. However, our objective in this effort is different from NCM; our focus lies in the causal effects pertaining to an NN model, primarily designed to enhance predictive accuracy. Our methodology remains applicable even when only partial knowledge of the underlying causal graph is accessible.

\noindent \textbf{Explainability:} In addition to promoting transparency in decision-making processes, the elucidation of NN models serves several purposes, including the identification of concealed biases present in data~\citep{alvarez2017causal}, the revelation of fairness~\citep{survey}, the debugging~\citep{lmdebugger} and enhancement of models through explanation-based regularizers~\citep{ross2017right, rieger2019interpretations,credo}. Numerous existing methods for explaining NN models quantify the impact of input features on model outputs using saliency maps~\citep{zeiler2014visualizing,simonyan2013deep,selvaraju2016grad}, local model approximations~\citep{ribeiro2016should}, approximations of output gradients with respect to inputs~\citep{sundararajan2017axiomatic,smilkov2017smoothgrad}, Shapley values~\citep{shapley,causalshapley}, among others.
In this work, we focus on the causal effects of input features on  output in an NN model, which can be very useful in safety-critical domains such as healthcare, aerospace, law, and defense. 
See Appendix \S\ref{unique_causal} to understand further why causal explanations are important through a real-world example.

\noindent \textbf{Causal Explanations:}
By considering an NN as an SCM, assuming that input features are $d$-separated from each other,~\citep{ca} proposed a post-hoc causal explanation method to find the average causal effects (ACE) in a trained NN. However, the assumption of independence among inputs limits their ability to consider indirect causal effects. Subsequent studies by~\citep{khademi2020causal,yadu2021icip,wang2021contrastive,cxplain,goyal2019explaining} have followed ACE as defined therein to quantify the learned causal effects. Other causal explanation methods utilize counterfactuals to analyze model behavior under semantically meaningful changes applied to inputs~\citep{surveyrecourse,goyal2019counterfactual,wachter,multiobjective,concepts,mothilal2021towards,mahajan2019preserving}. However, these methods are commonly employed for qualitative analysis of the model rather than computing causal effects.

\noindent \textbf{Direct and Indirect Explanations:}
Among existing efforts that explicitly investigate interactions among input variables while computing explanations for NN models, prominent methods are those based on Shapley values~\citep{shapley}. For instance, in the context of handling missing features in Shapley explanations, it is discouraged to sample from the conditional distribution (rather than the marginal distribution) because the inputs are independent with respect to the causal graph of the NN~\citep{acausalproblem}. While~\citep{causalshapley} considers both direct and indirect effects motivated by the direct and indirect pathways in the underlying causal graph, even if input neurons of the NN model being explained do not have causal connections, its focus is on providing Shapley values that may not necessarily be causal effects obtained from the adjustment formula (see Appendix \S\ref{sec cshap comparison}). In this study, we consider input feature interactions while learning and explaining causal effects in NNs. Our approach explicitly estimates and preserves indirect causal effects in an NN model. While~\citep{credo} discusses direct and total causal effects for NN model explanations, it does not focus on indirect causal effects. The work most closely related to ours is presented in~\citep{genderbias}, which examined both direct and indirect causal effects in Transformer-based language models for capturing gender bias. However, that study conducted a post-hoc analysis of such models for a different objective, whereas our proposed method represents an ante-hoc approach to learning and explaining both direct and indirect causal effects. Other related work, such as concept-based explanation methods, are discussed in Appendix \S\ref{sec related work concept based}.

\section{Causal Effects in Neural Networks}
\label{Section: causal effects in NNs}

Let $\mathcal{G}=(\mathbf{V},\mathbf{E})$ be a causal graph where $\mathbf{V}$ $=\{X_1, X_2, \dots, X_n, Y\}$ is the set of random variables and $\mathbf{E}$ is the set of edges denoting the causal influences among the variables in $\mathbf{V}$. 
Let $\mathbf{X}= \{X_1,\dots,X_n\}=\mathbf{V}\setminus\{Y\}$, $ch(X_i)=\{X_j|X_i\rightarrow X_j\}\subseteq \mathbf{V}\setminus \{X_i,Y\}$ be the set of children of $X_i$ except $Y$, and $pa(X_i)=\{X_j|X_i\leftarrow X_j\}\subseteq \mathbf{V}\setminus \{X_i, Y\}$ be the set of parents of $X_i$ except $Y$. This definition of $ch(X_i)$ and $pa(X_i)$ allows us to model indirect effects between input variables. 
Let $\mathcal{N}$ be an NN model that is trained to predict $Y$ given $\mathbf{X}$ as input by minimizing the empirical loss $\mathcal{R}_{ERM}$ in Eq~\ref{erm} for a given set $\mathcal{D}=\{(x_1^j,\dots,x_n^j, y^j)\}_{j=1}^N$ of $N$ data points.

\begin{equation}
\small
    \label{erm}
    \mathcal{R}_{ERM} = \displaystyle \frac{1}{N}\sum_{j=1}^N \mathcal{L}(y^j, \mathcal{N}(x_1^j,\dots,x_n^j))
\end{equation}

\noindent where $\mathcal{L}$ is an appropriate loss function such as root mean squared error for regression and cross-entropy loss for classification. Let $\hat{Y}=\mathcal{N}(X_1,\dots,X_n)$ be the overall output of the final layer of $\mathcal{N}$. $\mathcal{N}$ can be conceptualized as a directed acyclic graph (DAG) comprising directed edges connecting successive layers of neurons. Consequently, the output $\hat{Y}$ can be understood as the outcome arising from a series of interactions from the first to the final layer. When studying the causal effects of inputs on the output of $\mathcal{N}$, solely the neurons in the first and final layers are considered. Consequently, similar to~\citep{ca}, we can marginalize the influence of hidden layers within $\mathcal{N}$ and focus solely on the causal structure involving inputs and outputs (see Fig~\ref{fig:introduction} (a)). Note that while we follow~\citep{ca} in our view of NN as an SCM, they do not consider or model indirect effects, which is the focus of our work. 
To this end, we begin by defining various causal effects of input features on the output of a trained NN model.
\begin{definition}\textbf{(Average Causal Effect in an NN)}
\label{def: ACE}
The Average Causal Effect (ACE) of an input feature $X_i$ at an intervention $x_i$ with respect to a baseline intervention $x_i^*$ on the output $\hat{Y}$ of an NN $\mathcal{N}$ is defined as 
\begin{equation*}
\label{eq ace}
\small
{
 ACE^{\hat{Y}}_{X_i} = \mathbb{E}[\hat{Y}|do(X_i = 
x_i)] - \mathbb{E}[\hat{Y}|do(X_i = x_i^*)]   
}
\end{equation*}
\end{definition}
\noindent where $do(X_i = x_i)$ denotes an external intervention to the variable $X_i$ with the value $x_i$ (see Defn.~\ref{interventionaldistribution} in Appendix ~\ref{sec preliminaries}). We use $do(X_i)$ to refer to $do(X_i=x_i)$ when there is no ambiguity. $ACE$ is also called the average total causal effect, which is the sum of direct and indirect causal effects.
\begin{definition}
\label{def:direct_effect}
\textbf{(Average Direct Causal Effect in an NN)} The Average Direct Causal Effect (ADCE) measures the causal effect of a feature $X_i$ on the output $\hat{Y}$ of an NN when $\mathbf{Z}=ch(X_i)$ are intervened with values under the baseline intervention $do(X_i=x_i^*)$, denoted by $\mathbf{Z}_{X_i^*}$.
\begin{equation*}
\small
{
ADCE^{\hat{Y}}_{X_i} = \mathbb{E}[\hat{Y}|do(X_i,\mathbf{Z}_{X_i^*})]-\mathbb{E}[\hat{Y}|do(X_i^*, \mathbf{Z}_{X_i^*})]    
}
\end{equation*}
\end{definition}
\begin{definition}
\label{def:indirect_effects}
\textbf{(Average Indirect Causal Effect in an NN)} The Average Indirect Causal Effect (AICE) measures the causal effect of a feature $X_i$ on the output $\hat{Y}$ of an NN when $\mathbf{Z}=ch(X_i)$ are intervened with values under $do(X_i=x_i)$, denoted by $\mathbf{Z}_{X_i}$, while keeping the $X_i$ value fixed at the baseline intervention $do(X_i=x_i^*)$.
\begin{equation*}
\small
{
 AICE^{\hat{Y}}_{X_i} = \mathbb{E}[\hat{Y}|do(X_i^*,\mathbf{Z}_{X_i})]-\mathbb{E}[\hat{Y}|do(X_i^*, \mathbf{Z}_{X_i^*})]   
}
\end{equation*}
\end{definition}

\section{Learning and Explaining Direct and Indirect Causal Effects in Neural Networks}

We now present our methodology for learning and explaining indirect causal effects within NNs. 
Following~\citep{tarnet,dr_net,disent}, we make the following assumption concerning the underlying causal graph $\mathcal{G}$.
\begin{assumption}
\label{assumption1}
There are no latent (unobserved) confounders in the underlying causal graph $\mathcal{G}$. 
\end{assumption}

To quantify direct and indirect causal effects of an input $X_i$ on the output $\hat{Y}$ of an NN, it is required to perform an intervention on $ch(X_i)$ with specific values based on $X_i$'s value (as formally stated in Defns~\ref{def:direct_effect} and \ref{def:indirect_effects}). The above assumption allows us to get the values to perform an intervention on $ch(X_i)$. 
\begin{hypothesis}
\label{hypothesis1}
In an NN $\mathcal{N}$, the indirect effect of a variable $X_i$ on $Y$ via $ch(X_i)$, $AICE_{X_i}^{\hat{Y}}$, is identifiable in $\mathcal{N}$ iff there are feedforward edges from $X_i$ to $ch(X_i)$ in the architecture of $\mathcal{N}$.
\end{hypothesis}

The supporting proof for the above hypothesis is straightforward and provided in Appendix \S\ref{sec identifiability}. Note that the edges between $X_i$ and $ch(X_i)$ capture the true causal relationships in the real-world. In such an architecture of $\mathcal{N}$ with lateral edges between $X_i$ and $ch(X_i)$, the weights parametrizing these edges are also learned by $\mathcal{N}$ along with other weights in the model while optimizing for $\mathcal{N}$'s objective.

Although Hypothesis~\ref{hypothesis1} may appear self-evident, it has been overlooked in existing methods for explaining NN models. For example,~\citep{acausalproblem} argue that  \textit{Shapley} explanations in a simple feedforward NN should treat all input features to be independent because the causal graph of a simple feedforward NN has no causal connections among input neurons. A similar argument is given by~\citep{Datta2016AlgorithmicTV} focusing on only \textit{direct} effects while quantifying input influence on the output of an NN. Not accounting for indirect effects when modeling statistical relationships in the observed data distribution (e.g., using conditional expectation instead of marginal expectation for missing features while calculating Shapley values) may generate incorrect explanations~\citep{acausalproblem}. 


\subsection{Learning Indirect Causal Effects}
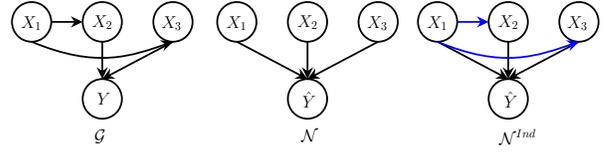
\begin{figure}
    \centering
    \scalebox{0.45}{
    \tikzset{every picture/.style={line width=0.75pt}}  
    \input{images/ahce.tikz}
    }
    
     \caption{\footnotesize Comparison of the proposed architecture $\mathcal{N}^{Ind}$ with a traditional NN architecture $\mathcal{N}$. $\mathcal{G}$ is the ground truth causal graph. $\mathcal{N}$ and $\mathcal{N}^{Ind}$ differ in input layer such that the inputs in $\mathcal{N}^{Ind}$ are connected (shown in blue color) according to the causal edges in $\mathcal{G}$. In contrast, the inputs in $\mathcal{N}$ are independent. $\mathcal{N}$ and $\mathcal{N}^{Ind}$ may contain edges that are not present in $\mathcal{G}$ due to the feedforward connections from input layer to predictions in NN architecture (e.g., $X_1\rightarrow \hat{Y}$ is present in $\mathcal{N}$, $\mathcal{N}^{Ind}$ but not in $\mathcal{G}$).}
    \label{fig:intvserm}
\end{figure}

Following the above discussion, given a standard NN $\mathcal{N}$, we propose an augmented NN architecture $\mathcal{N}^{Ind}$ for capturing indirect causal effects of input features on the output. $\mathcal{N}^{Ind}$ contains lateral directed connections among the input neurons based on the available knowledge of the true causal graph (see Fig~\ref{fig:intvserm}). Our methodology remains applicable even when only a partial causal graph is available, capturing indirect effects exclusively on the available connections. We call the set of NN edges introduced among input neurons as layer $0$ connections to separate them from NN connections in hidden layers. These connections among input features have learnable parameters akin to other parameters within the NN. 
\begin{algorithm}
\footnotesize
\caption{Pseudocode for training $\mathcal{N}^{Ind}$ model}
\label{algo:nn_edges_input_layer}
\begin{algorithmic}[1]
   \STATE {\bfseries Input:} True causal graph $\mathcal{G}$, $\mathcal{D} = \{(x^j_1,\dots,x^j_n,y^j)\}_{j=1}^{N}$,  parameters $\theta_0,\dots,\theta_m$ of layers $l_0,\dots,l_m$ of $\mathcal{N}^{Ind}$, $\lambda$, functions $f_i^0$ in $l_0$ learned by introducing edges among input features. 
   \STATE {\bfseries Output:} Trained $\mathcal{N}^{Ind}$ model  
\FOR{each epoch}
\FOR{phase in $[1,2]$}
\IF{phase = $1$}
\STATE {\scriptsize $\mathcal{R}_{ERM} = \displaystyle \frac{1}{N}\sum_{j=1}^N \mathcal{L}(y^j, \mathcal{N}^{Ind}(x_1^j,\dots,x_n^j))$}
\STATE {Compute gradients of $\mathcal{R}_{ERM}$ w.r.t. $\theta_1,\dots,\theta_m$}
\STATE {Update the parameters $\theta_1,\dots,\theta_m$ using SGD}
\ELSE 

\FOR{ each $(x^j_1,\dots,x^j_n,y^j)$ in $\mathcal{D}$}
\STATE{$x_i^j = f_i^0(pa(x_i^j))\ \forall i\ s.t.\ pa(X_i)\neq \emptyset $}
\ENDFOR

\STATE {\scriptsize $\mathcal{R} = \mathcal{R}_{ERM}+\displaystyle \lambda \sum_{j=1}^N \sum_{\{\forall X_i|pa(X_i)\neq \emptyset\}} (x^j_i-f^0_i(pa(x^j_i)))^2$}
\STATE {Compute gradients of $\mathcal{R}$ w.r.t. $\theta_0,\dots,\theta_m$}
\STATE {Update parameters of $\theta_0,\dots,\theta_m$ using SGD}
\ENDIF
\ENDFOR
\ENDFOR
\STATE{return trained $\mathcal{N}^{Ind}$}
\end{algorithmic}
\end{algorithm}

To train the augmented $\mathcal{N}^{Ind}$ model, we propose an ante-hoc training algorithm consisting of two phases, each of which is invoked sequentially in each epoch. In the first phase, we freeze the parameters of the layer $0$ and train the remaining part of the NN. In the second phase, we train the entire model i.e., parameters of layer $0$ to the final layer. In the second phase, the input to the $\mathcal{N}^{Ind}$ model is constructed as follows. Consider a specific input data point $(x^j_1,\dots,x^j_n) \sim \mathcal{D}$. The value of each input variable $X_i$ for which $pa(X_i)=\emptyset$ is taken from $(x^j_1,\dots,x^j_n)$, and the remaining input feature values are \textit{derived topologically} by feeding the other input variables into layer $0$. That is, for each $X_i$ with $pa(X_i)\neq \emptyset$, if $f_i^0$ is the function of its parents $pa(X_i)$ in layer $0$, we derive $X_i = f_i^0(pa(X_i))$. Please note that $f_i^0$ is modeled by the NN connections in layer $0$. These two training phases are carried out sequentially in every epoch until we reach the desired minimum loss value (or appropriate stopping condition). To aid better learning of parameters of layer 0, we add a regularization term to the empirical loss $\mathcal{R}_{ERM}$ in Eqn~\ref{erm} that incurs a penalty if the derived feature values deviate from actual feature values in the training data. Eqn~\ref{regularizer} shows the overall loss value used in phase 2 with regularization term and corresponding regularization hyperparameter $\lambda$. $\mathcal{N}^{Ind}$ is trained using stochastic gradient descent (SGD), as with any other NN model. Algorithm~\ref{algo:nn_edges_input_layer} summarizes this training procedure.

\begin{equation}
\footnotesize
    \label{regularizer}
    \mathcal{R} = \mathcal{R}_{ERM}+\displaystyle \lambda \sum_{j=1}^N \sum_{\{\forall i|pa(X_i)\neq \emptyset\}} (x^j_i-f_i^0(pa(x^j_i)))^2
\end{equation}

\subsection{Explaining Indirect Causal Effects}
On training the ante-hoc model $\mathcal{N}^{Ind}$, we now present a methodology to compute the acquired indirect causal effects in the learned model. We begin by formally defining causal effect \textit{identifiability} in this context. 

\begin{definition}\textbf{Causal Effect Identifiability in an NN.} The causal effect of an input feature $X_i$ on the output $\hat{Y}$ of an NN is identifiable if $p(\hat{Y}|do(X_i))$ can be computed uniquely from any positive probability distribution $p(X_1,\dots,X_n,\hat{Y})$.
\end{definition}

Under the \textit{no latent confounding} assumption (Assumption~\ref{assumption1}), following Theorem 3.2.5 and Corollary 3.2.6 of~\citep{pearl2009causality}, it is easy to show that $ADCE^{\hat{Y}}_{X_i}$ and $AICE^{\hat{Y}}_{X_i}$ are identifiable in $\mathcal{N}^{Ind}$ (we provide formal proofs in Appendix \S\ref{sec identifiability}). Now, to evaluate $ADCE^{\hat{Y}}_{X_i}$ and $AICE^{\hat{Y}}_{X_i}$ in $\mathcal{N}^{Ind}$, we need to in turn evaluate the following quantities: $\mathbb{E}[\hat{Y}|do(X_i^*, \mathbf{Z}_{X_i^*})]$, $\mathbb{E}[\hat{Y}|do(X_i, \mathbf{Z}_{X_i^*})] \text{ and } \mathbb{E}[\hat{Y}|do(X_i^*, \mathbf{Z}_{X_i})]$ (see Defns~\ref{def:direct_effect}, \ref{def:indirect_effects} and recall that $\mathbf{Z}=ch(X_i)$). These terms, which are of the form $\mathbb{E}[\hat{Y}|do(\mathbf{S})]$ where $\mathbf{S}$ is a set of features, often require us to marginalize over other input features $\mathbf{X}\setminus \mathbf{S}$ as:
\begin{equation}
\mathbb{E}[\hat{Y}|do(\mathbf{S})] = \mathbb{E}_{\mathbf{X}\setminus \mathbf{S}}\left[\mathbb{E}[\hat{Y}|\mathbf{S}, \mathbf{X}\setminus \mathbf{S}]\right]
\label{interventional eq}
\end{equation}
Evaluating the above expression, typically using an \textit{adjustment set} (see Defn~\ref{def:validadjustment} in Appendix \S\ref{sec preliminaries}), can incur significant computational overhead, which grows exponentially with the number of features in $\mathbf{X}\setminus \mathbf{S}$, especially when they are continuous and real-valued. To avoid such prohibitive computational requirements, following earlier work~\citep{montavon2017explaining,ca}, we consider the second-order Taylor's approximation to the NN output $\hat{Y}=f(\mathbf{X})$ 
around the mean vector $\mu$, where $\mu_j = \mathbb{E}[X_j|do(\mathbf{S})]$ as follows:
\begin{equation*}
\small
{
\begin{aligned}
f(\mathbf{X}) &\approx f(\mu)  +\\
&\nabla^T f(\mu)(\mathbf{X} - \mu) + \frac{1}{2}(\mathbf{X} - \mu)^T\nabla^2f(\mu)(\mathbf{X} - \mu)
\end{aligned}
}
\label{taylor expansion eq}
\end{equation*}
Taking interventional expectations on both sides gives:
\begin{equation}
\small
{
\begin{aligned}
\mathbb{E}[f(\mathbf{X})| do(\mathbf{S})] &\approx f(\mu) + \\
&\frac{1}{2}Tr(\nabla^2f(\mu)\mathbb{E}[(\mathbf{X} - \mu)(\mathbf{X} - \mu)^T| do(\mathbf{S})])
\end{aligned}
}
\label{expected taylor expansion eq}
\end{equation}
The first-order terms vanish because $\mathbb{E}[\mathbf{X}|do(\mathbf{S})] = \mu$. To evaluate Eqn~\ref{expected taylor expansion eq}, we need to calculate the interventional mean vector $\mu = \mathbb{E}[\mathbf{X}|do(\mathbf{S})]$ and the interventional covariance matrix $\mathbb{E}[(\mathbf{X} - \mu)(\mathbf{X} - \mu)^T| do(\mathbf{S})]$.

We present the following steps $S1-S4$ to evaluate interventional means and covariances for interventions: $do(X_i^*, \mathbf{Z}_{X_i^*}), do(X_i, \mathbf{Z}_{X_i^*})$, and $do(X_i^*, \mathbf{Z}_{X_i})$.

\begin{itemize}
\item [S1.] For an intervention on $X_i$ with the value $x_i$, set $\mu[i]=x_i$.

\item [S2.] To get interventional values $\mathbf{Z}_{X_i}$ for the variables in $\mathbf{Z}$ under the intervention $do(X_i=x_i)$, for each variable $X_p\in \mathbf{Z}$ taken in topological order, compute $X_p = f_0^p(pa(X_p))$ and $\mu[p] = \mathbb{E}_{pa(X_p)}\big[\mathbb{E}[X_p|X_i,pa(X_p)\setminus \{X_i\}]\big]$. This step accounts for updating the values of children of $X_i$ based on the intervention on $X_i$.

\item [S3.]  For each variable $X_q\not \in \mathbf{Z}$, set $\mu[q] = \mathbb{E}[X_q]$.

\item [S4.] Compute the interventional covariance matrix from the interventional data distribution obtained after performing step $S2$.
\end{itemize}

After performing the above steps, we can substitute the interventional mean and covariance matrix in Eqn~\ref{expected taylor expansion eq} to evaluate the expressions $\mathbb{E}[\hat{Y}|do(X_i^*, \mathbf{Z}_{X_i^*})]$, $\mathbb{E}[\hat{Y}|do(X_i, \mathbf{Z}_{X_i^*})], \mathbb{E}[\hat{Y}|do(X_i^*, \mathbf{Z}_{X_i})]$. An algorithm summarizing this overall procedure of evaluating $ADCE^{\hat{Y}}_{X_i}$, $AICE^{\hat{Y}}_{X_i}$ in $\mathcal{N}^{Ind}$ is provided in Appendix \S~\ref{algorithms}.

\subsection{Efficient Implementation Strategies}
\label{sec:improve_time_and_space_complexity}
Computation of causal effects, in general, can be compute and memory intensive. We hence also provide a few efficient implementation strategies for such computations, which we also incorporate in our experiments.
Let each input $X_i\in \mathbf{X}$ assume one of $k$ possible values ($k=2$ in the binary case). Evaluating causal effects takes roughly $\mathcal{O}(n^k)$ time because of the marginalization step in Eqn~\ref{interventional eq}, where $n$ is the dimensionality of the input vector $\mathbf{X}$. Evaluating the approximation in Eqn~\ref{expected taylor expansion eq} also scales in the order of $\mathcal{O}(n^2)$ as an input intervention may affect all children (Defns~\ref{def:direct_effect},~\ref{def:indirect_effects}). These limitations get accentuated in architectures such as Recurrent Neural Networks (RNNs)  (see Appendix~\ref{implementation} for complexity analysis in RNNs). To address these issues, we propose the following improvements. 

\noindent \textbf{Runtime Efficiency using Binning:}
Computing causal effects using Eqn~\ref{expected taylor expansion eq} requires computing the interventional mean and interventional covariance (interventional statistics).  To speed up this calculation, we divide the computation into offline and online phases. In the offline step (which can be done independent of the NN training phase), for every data point $\mathbf{X}$ in the training set, we generate and store the interventional statistics for all features $X_i\in \mathbf{X}$ for all interventional values. In the online phase, to find the causal effect for feature $X_i$ with intervention value of $x_i$ in a test data point $\mathbf{X}_{te}$, we first find the data point $\mathbf{X}_{tr}$ in the training set that is most similar to $\mathbf{X}_{te}$. Let $\alpha$ be the value taken by feature $X_i$ in $\mathbf{X}_{tr}$, closest to $X_i$. We access the interventional statistics stored for $\mathbf{X}_{tr}$ corresponding to feature $X_i$ with intervention $\alpha$ (computed in the offline phase). 
This retrieved \textit{nearest} interventional statistics is used for causal effect computation. 
This procedure, detailed further in Appendix \S~\ref{implementation}, reduces significant runtime leveraging offline computations. We refer to this approach as \textit{binning} since a training sample captures a bin and acts as a proxy for other samples/values in its neighborhood. To further speed up ACE computation, we exploit the fact that the Hessian term $\nabla^2f$ in Eqn~\ref{expected taylor expansion eq} can be approximated using $J^TJ$ where $J$ is the Jacobian of the NN model function (using Gauss-Newton Hessian approximation). 

\noindent \textbf{Memory Requirements:}
Storing offline interventional statistics for every sample on the dataset (and corresponding intervention values) quickly becomes impractical, especially for high-dimensional data.  
To reduce this memory overhead, we use clustering/hashing techniques (KD Tree, DBSCAN) to cluster training data samples, and store interventional statistics for only cluster centers (see Appendix \S\ref{implementation} for more details of this strategy). From the results shown in Tab~\ref{tab:total_results}, Appendix \S\ref{implementation}, we observe 3 to 10-fold improvements in run time using the proposed binning approach for a slight reduction in the precision of estimated causal effects.

\section{Experiments and Results}
\label{sec experiments}
We conduct a wide range of experiments across three kinds of datasets -- A synthetic dataset, three well-known public real-world benchmark datasets, and three industry-based simulated datasets. We compare the causal explanations of AHCE with a well-known post-hoc gradient-based explanation method: Integrated Gradients (IG)~\citep{sundararajan2017axiomatic}, a post-hoc causal explanations (CA) method~\citep{ca}, the causal Shapley values (CSHAP)~\citep{causalshapley}, and a recent causal regularization method in~\citep{credo}. We compare against IG since it can be viewed as computing individual causal effects of input neurons on the output of an NN~\citep{imbens_rubin_2015}. Ground truth causal effects are computed using the adjustment formula in Eqn~\ref{interventional eq}. Following~\citep{credo}, we use the Root Mean Squared Error (RMSE) and Frechet distance between true causal effects and the learned explanations. We present our results on total causal effects for a fair comparison with all methods (indirect causal effects do not exist for IG, CA, CREDO; comparison of CSHAP with our method on indirect causal effects is presented in Appendix \S\ref{sec setup and additional results}). Additional results and experimental setup are presented in Appendix \S\ref{sec setup and additional results} owing to space constraints.  Our code is provided in the supplementary material.

\noindent \textbf{Synthetic Data:}
\label{synthetic_tablular_1_expts}
We create a synthetic dataset using a causal graph and corresponding structural equations shown in Fig~\ref{fig synthetic data} and Tab~\ref{tab synthetic data1}. From Fig~\ref{fig synthetic data}, $W$ has only indirect causal effect on $Y$ via the paths: $W\rightarrow X\rightarrow Y, W\rightarrow Z\rightarrow X \rightarrow Y$, and $W\rightarrow Z \rightarrow Y$. $Z$ has a direct causal effect on $Y$ via the path $Z\rightarrow Y$ and an indirect causal effect on $Y$ via the path $Z\rightarrow X\rightarrow Y$, and $X$ has only a direct causal effect on $Y$ via the path $X\rightarrow Y$. This dataset has linear equations with additive Gaussian noise among input features $W,Z,X$, and the output $Y$ is a non-linear function of its inputs with additive Gaussian noise. Hence, for purposes of modeling causal effects, the lateral connections among inputs in $\mathcal{N}^{Ind}$ are obtained using simple linear regressors that take a set of real numbers as input and produce a real number as output (for real-world datasets, we replace simple linear regressors with multi-layer perceptrons to account for non-linear relationships among input features). 

\begin{figure}[H]

  \begin{minipage}[b]{0.5\linewidth}
    \scalebox{0.47}{
    \tikzset{every picture/.style={line width=0.75pt}}  
    \input{images/synthetic.tikz}
    }
    \captionof{figure}{\footnotesize Synthetic DAG}
    \label{fig synthetic data}
  \end{minipage}%
  \scalebox{0.9}{
  \begin{minipage}[b]{0.5\linewidth}
    \centering
    \small
     \begin{align*}
     W &\leftarrow Uniform(0, 1)\\
     Z &\leftarrow W/2 + \mathcal{N}(0, 0.1)\\
     X &\leftarrow -W-Z + \mathcal{N}(0, 0.1)\\
     Y &\leftarrow X^3 + \log(Z^2) + \mathcal{N}(0, 0.1)
    \end{align*}
    \captionof{table}{Synthetic Equations}
        \label{tab synthetic data1}
  \end{minipage}%
  }
  
\end{figure}

Tab~\ref{tab:synthetic_results} shows the results. The total causal effects given by our method are closer to ground truth causal effects compared to baselines. These results show that the training algorithm for our ante-hoc causal explanation model can better learn both direct and indirect causal effects and match the total causal effects.

We perform an ablation study to empirically verify our hypothesis that adding edges among input features helps in learning causal effects. In this study, we check the causal effects with and without outgoing edges from each input feature $W, Z, X$ in $\mathcal{N}^{Ind}$ (denoted by $\mathcal{N}^{Ind}_{With}$, $\mathcal{N}^{Ind}_{Without}$ respectively in Tab~\ref{tab:synthetic_ablation}). This setting also resembles real-world scenarios where we can often access only a partial causal graph. From the results in Tab~\ref{tab:synthetic_ablation}, we observe that the total causal effects in $\mathcal{N}^{Ind}_{With}$ are closer to the ground truth total causal effects compared to $\mathcal{N}^{Ind}_{Without}$.

\begin{table}[H]
      
      \footnotesize
      \centering
     \scalebox{0.75}{
             \begin{tabular}{ll|cccc|c}
                \toprule
                &\textbf{Feature}&\textbf{IG}& \textbf{CA} 
                 &\textbf{CSHAP}& \textbf{CREDO}& \textbf{AHCE}\\
&&\citepalias{sundararajan2017axiomatic}&\citepalias{ca}&\citepalias{causalshapley}&\citepalias{credo}&\textbf{(Ours)}\\
                \midrule
                &W &0.10$\pm$0.00&0.09$\pm$0.00&0.24$\pm$0.00&0.08$\pm$0.01&0.04$\pm$0.02\\
                &Z &0.11$\pm$0.04&0.04$\pm$0.01&0.30$\pm$0.00&0.06$\pm$0.00&0.05$\pm$0.00\\
                &X &0.12$\pm$0.00&0.11$\pm$0.00&0.25$\pm$0.01&0.11$\pm$0.00&0.10$\pm$0.02\\
                \cmidrule{2-7}
                \parbox[t]{2mm}{\multirow{-4}{*}{\rotatebox[origin=c]{90}{RMSE ($\downarrow$)}}}&Average &0.11$\pm$0.02&0.08$\pm$0.00&0.26$\pm$ 0.00&0.08$\pm$0.01&\cellcolor{gray!50}\textbf{0.06$\pm$0.01}\\
                \midrule
                &W &0.25$\pm$0.00&0.25$\pm$0.00&0.25$\pm$ 0.05&0.23$\pm$0.03&0.14$\pm$0.06\\
                &Z &0.19$\pm$0.05&0.09$\pm$0.05&0.33$\pm$ 0.02&0.16$\pm$0.01&0.13$\pm$0.02\\
                &X &0.24$\pm$0.07&0.23$\pm$0.04&0.32$\pm$ 0.04&0.26$\pm$0.03&0.24$\pm$0.03\\
                \cmidrule{2-7}
                \parbox[t]{2mm}{\multirow{-4}{*}{\rotatebox[origin=c]{90}{Frechet ($\downarrow$)}}}&Average &0.23$\pm$0.04&0.19$\pm$0.03&0.30$\pm$ 0.04&0.22$\pm$0.02&\cellcolor{gray!50}\textbf{0.17$\pm$0.04}\\
                \bottomrule
             \end{tabular}
             }
             
             \caption{\footnotesize Results on synthetic dataset.}
             \label{tab:synthetic_results}
             
    \end{table}
    
\begin{table}[H]
\footnotesize

    \centering
     \scalebox{0.8}{
             \begin{tabular}{l|cc|cc}
                \toprule
                Feature& \multicolumn{2}{c|}{RMSE ($\downarrow$)}  & \multicolumn{2}{c}{Frechet ($\downarrow$)}\\
                \midrule
                &$\mathcal{N}^{Ind}_{Without}$& $\mathcal{N}^{Ind}_{With}$& $\mathcal{N}^{Ind}_{Without}$&$\mathcal{N}^{Ind}_{With}$\\
                \midrule
                W&0.08$\pm$0.01&0.04$\pm$0.02&0.23$\pm$0.02&0.14$\pm$0.06\\
                Z&0.06$\pm$0.01&0.05$\pm$0.00&0.15$\pm$0.02&0.13$\pm$0.02\\
                X&0.10$\pm$0.02&0.10$\pm$0.02&0.24$\pm$0.03&0.24$\pm$0.03\\
        \cmidrule{1-5}
        Average&0.08$\pm$0.01&\cellcolor{gray!50}\textbf{0.06$\pm$0.01}&0.20$\pm$0.02&\cellcolor{gray!50}\textbf{0.17$\pm$0.04}\\
            \bottomrule
             \end{tabular}
             }
             
             \captionof{table}{
             Ablation study on synthetic dataset.}
             \label{tab:synthetic_ablation}
             
    \end{table}

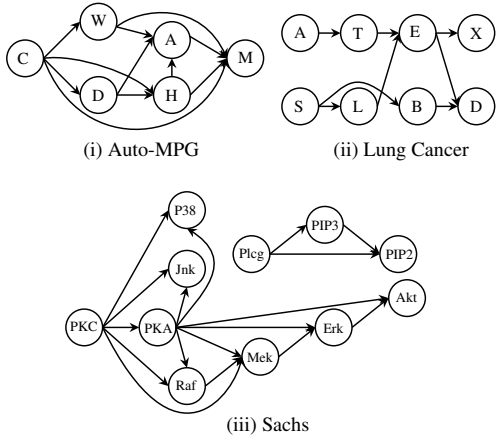
\begin{figure}
    \centering
    \scalebox{0.37}{
    \tikzset{every picture/.style={line width=0.75pt}}  
    \input{images/autompg_asia.tikz}
    }
    \caption{\footnotesize (i) Estimated causal graph of Auto-MPG dataset where C: number of cylinders, D: displacement, W: weight, H: horsepower, A: acceleration, and M: miles per gallon. (ii) True causal graph of lung cancer dataset where A: visit to Asia, T: tuberculosis, S: smoking, L: lung cancer, B: bronchitis, E: either T or L, X: X-ray, and D: dyspnea. (iii) True causal graph of Sachs dataset.}
    \label{fig autompg asia sachs}
    
\end{figure}

\noindent \textbf{Auto-MPG}: 
In this experiment, we work on Auto-MPG dataset~\citep{Dua2019} where the task is to predict \textit{miles per gallon} (MPG) based on various parameters such as \textit{acceleration, horsepower}, etc. We do not know the ground truth causal graph in this case. Hence, we first construct a causal graph as shown in Fig~\ref{fig autompg asia sachs} based on pertinent domain knowledge. Subsequently, we verify the correctness of this constructed causal graph through interaction with the popular large language model GPT-3.5~\citep{brown2020language}, questioning the correctness of each causal edge within the constructed graph. We use this constructed graph as the available knowledge in our experiments. Tab~\ref{tab:cancerandautompg} shows these results. Since the underlying structural equations are unavailable for this dataset, we cannot evaluate indirect causal effects. However, we can compare the performance with respect to total causal effects, which is the sum of direct and indirect causal effects. From the results, our method outperforms baselines in capturing true total causal effects. 

\noindent \textbf{Lung Cancer:}
In Lung Cancer dataset~\citep{scutari2014bayesian}, whose causal graph is known (see Fig~\ref{fig autompg asia sachs}), we consider \textit{Dyspnea} is the output variable with the remaining features such as \textit{smoking, bronchitis}, etc. as inputs. From the results shown in Tab~\ref{tab:cancerandautompg}, our model is better at learning the true total causal effects when compared to the baselines. The lateral connections among input features are implemented using simple multi-layer perceptrons with non-linear activation functions. Since the underlying causal graph of the Lung Cancer dataset is a discrete Bayesian network with binary-valued features, the Frechet score is not relevant, and so we report only RMSE values for this dataset. Similar to Auto-MPG dataset, we present results on total causal effects.
\noindent\textbf{Sachs:} Sachs dataset consists of 11 protein types and their causal relationships as shown in Fig~\ref{fig autompg asia sachs}. We consider the variable \textit{Akt} as output and the remaining variables \textit{Plcg, PIP2, PIP3} as inputs as they do not causally influence the output.
\begin{table}
    \centering
    
    \footnotesize
    \scalebox{0.73}{
    \begin{tabular}{ll|cccc|c}
         \toprule
    &\textbf{Feature}&\textbf{IG}&\textbf{CA}&\textbf{CSHAP}&\textbf{CREDO}&\textbf{AHCE}\\
&&\citepalias{sundararajan2017axiomatic}&\citepalias{ca}&\citepalias{causalshapley}&\citepalias{credo}&(Ours)\\
        \midrule
        \multicolumn{7}{c}{\textbf{Auto-MPG}}  \\
        \midrule
         &Cylinders&0.12$\pm$0.00&0.13$\pm$0.00&0.20$\pm$0.00&0.11$\pm$0.02&0.01$\pm$0.00\\
         &Disp.&0.11$\pm$0.00&0.11$\pm$0.00&0.20$\pm$0.00&0.09$\pm$0.02&0.11$\pm$0.01\\
         &Horsepow.&0.21$\pm$0.02&0.04$\pm$0.01&0.17$\pm$0.00&0.07$\pm$0.02&0.09$\pm$0.01\\
         &Weight&0.27$\pm$0.04&0.09$\pm$0.00&0.05$\pm$0.00&0.09$\pm$0.02&0.07$\pm$0.00\\
         &Acceler.&0.07$\pm$0.01&0.07$\pm$0.00&0.02$\pm$0.00&0.15$\pm$0.05&0.07$\pm$0.00\\
         \cmidrule{2-7}
     \parbox[t]{2mm}{\multirow{-6}{*}{\rotatebox[origin=c]{90}{RMSE ($\downarrow$)}}}&Average&0.16$\pm$0.02&0.09$\pm$0.00&0.13$\pm$0.00&0.10$\pm$0.02&\cellcolor{gray!50}\textbf{0.07$\pm$0.00}\\
     \midrule
        &Cylinders&0.27$\pm$0.00&0.25$\pm$0.00&0.37$\pm$0.00&0.22$\pm$0.04&0.03$\pm$0.03\\
         &Disp.&0.25$\pm$0.00&0.21$\pm$0.01&0.38$\pm$0.00&0.19$\pm$0.03&0.21$\pm$0.02\\
         &Horsepow.&0.25$\pm$0.02&0.07$\pm$0.02&0.30$\pm$0.00&0.15$\pm$0.03&0.18$\pm$0.03\\
         &Weight&0.45$\pm$0.08&0.15$\pm$0.02&0.06$\pm$0.02&0.17$\pm$0.06&0.09$\pm$0.01\\
         &Acceler.&0.12$\pm$0.01&0.09$\pm$0.01&0.06$\pm$0.01&0.33$\pm$0.16&0.10$\pm$0.00\\
         \cmidrule{2-7}
     \parbox[t]{2mm}{\multirow{-6}{*}{\rotatebox[origin=c]{90}{Frechet ($\downarrow$)}}}&Average&0.27$\pm$0.02&0.16$\pm$0.01&0.23$\pm$0.00&0.21$\pm$0.06&\cellcolor{gray!50}\textbf{0.12$\pm$0.02}\\
        \midrule
        \multicolumn{7}{c}{\textbf{Lung Cancer}}  \\
        \midrule
        &Asia &0.46$\pm$0.05& 0.38$\pm$0.11&0.00$\pm$0.00&0.00$\pm$0.00&0.05$\pm$0.06\\
        &Tub. &0.62$\pm$0.04&1.13$\pm$0.04&0.99$\pm$0.00&1.00$\pm$0.00&0.58$\pm$0.29\\
        &Smoking &1.07$\pm$0.07& 1.01$\pm$0.00&0.99$\pm$ 0.00&1.00$\pm$0.00&0.56$\pm$0.33\\
        &L.Cancer &0.40$\pm$0.07&0.62$\pm$0.02&0.48$\pm$0.04&0.49$\pm$0.00&0.77$\pm$0.75\\
        &Bronch. &1.55$\pm$0.14&1.48$\pm$0.06&0.93$\pm$0.00&1.08$\pm$0.01&1.11$\pm$0.51\\
        &Either &0.87$\pm$0.18& 0.78$\pm$0.06&0.53$\pm$0.03&0.55$\pm$0.00&0.65$\pm$0.23\\
        &X-ray &0.11$\pm$0.05&0.09$\pm$0.04&0.03$\pm$0.00&0.00$\pm$0.00&0.08$\pm$0.12\\
        \cmidrule{2-7}
        \parbox[t]{2mm}{\multirow{-8}{*}{\rotatebox[origin=c]{90}{RMSE ($\downarrow$)}}}&Average&0.72$\pm$0.09&0.78$\pm$0.05&0.56$\pm$0.00&0.59$\pm$0.00&\cellcolor{gray!50}\textbf{0.54$\pm$0.33}\\
        \midrule
        \multicolumn{7}{c}{\textbf{Sachs}}  \\
        \midrule
         &PKC&0.08$\pm$0.07&0.10$\pm$0.09&0.19$\pm$0.00&0.08$\pm$0.02&0.12$\pm$0.06\\
         &PKA&2.29$\pm$1.40&2.19$\pm$0.90&0.46$\pm$0.00&3.81$\pm$0.02&0.65$\pm$0.17\\
         &Raf&0.15$\pm$0.03&0.11$\pm$0.05&0.24$\pm$0.00&0.02$\pm$0.02&0.12$\pm$0.03\\
         &Mek&0.20$\pm$0.04&0.21$\pm$0.13&0.23$\pm$0.00&0.42$\pm$0.02&0.14$\pm$0.01\\
         &Erk&4.33$\pm$3.25&0.63$\pm$2.23&0.53$\pm$0.00&2.87$\pm$0.05&0.51$\pm$0.34\\
         &Jnk&0.08$\pm$0.04&0.07$\pm$0.04&0.25$\pm$0.00&0.13$\pm$0.05&0.01$\pm$0.01\\
         &P38&0.26$\pm$0.18&0.09$\pm$0.06&0.31$\pm$0.00&0.04$\pm$0.05&0.02$\pm$0.01\\
         \cmidrule{2-7}
     \parbox[t]{2mm}{\multirow{-8}{*}{\rotatebox[origin=c]{90}{RMSE ($\downarrow$)}}}&Average&1.05$\pm$0.71&0.71$\pm$0.27&0.32$\pm$0.00&1.05$\pm$0.00&\cellcolor{gray!50}\textbf{0.22$\pm$0.09}\\
    \midrule
         &PKC&0.14$\pm$0.12&0.13$\pm$0.12&0.30$\pm$0.00&0.11$\pm$0.00&0.17$\pm$0.09\\
         &PKA&2.89$\pm$1.62&2.97$\pm$1.14&0.29$\pm$0.00&5.02$\pm$0.02&0.91$\pm$0.23\\
         &Raf&0.21$\pm$0.05&0.16$\pm$0.08&0.27$\pm$0.00&0.03$\pm$0.02&0.17$\pm$0.05\\
         &Mek&0.33$\pm$0.08&0.27$\pm$0.18&0.37$\pm$0.00&0.56$\pm$0.02&0.17$\pm$0.01\\
         &Erk&5.63$\pm$4.04&3.12$\pm$0.90&0.36$\pm$0.00&4.04$\pm$0.05&0.70$\pm$0.45\\
         &Jnk&0.12$\pm$0.06&0.09$\pm$0.05&0.36$\pm$0.00&0.16$\pm$0.05&0.02$\pm$0.02\\
         &P38&0.41$\pm$0.30&0.12$\pm$0.09&0.47$\pm$0.00&0.06$\pm$0.05&0.02$\pm$0.02\\
     \cmidrule{2-7}
     \parbox[t]{2mm}{\multirow{-8}{*}{\rotatebox[origin=c]{90}{Frechet ($\downarrow$)}}}&Average&1.39$\pm$0.90&0.98$\pm$0.36&0.34$\pm$0.00&1.43$\pm$0.00&\cellcolor{gray!50}\textbf{0.31$\pm$0.12}\\    \bottomrule
    \end{tabular}
    }
    \caption{\footnotesize Results on Auto-MPG, Lung Cancer, and Sachs datasets}
    \label{tab:cancerandautompg}
\end{table}
The results in Tab~\ref{tab:cancerandautompg} show that our model is better at learning the true total causal effects than the baselines.

\begin{figure}
    \centering
    \includegraphics[width=0.45\textwidth]{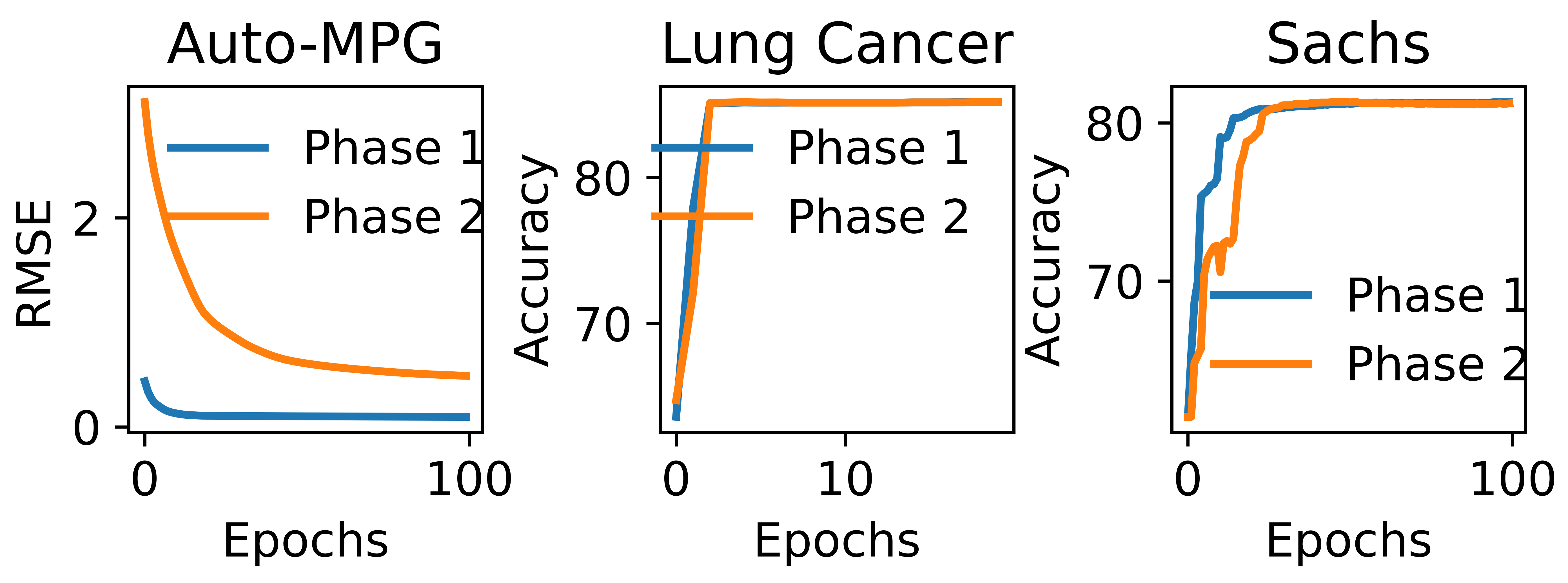}
    \caption{\footnotesize Convergence of RMSE/Accuracy values at the end of two phases of training outlined in Algorithm~\ref{algo:nn_edges_input_layer}.}
    \label{fig:convergence}
\end{figure}

To check that the second phase of our proposed ante-hoc training algorithm, where we train the parameters of layer $0$, does not reduce the model performance, we check the performance on each dataset after each epoch for each training phase. As shown in Fig~\ref{fig:convergence}, the two phases of ante-hoc model training converge without oscillations in phase 2 training.
\noindent \textbf{Flight Simulation Datasets:}
\label{exp:improvements}
To study the value of our efficient implementation strategies discussed in Sec.~\ref{sec:improve_time_and_space_complexity}, we consider flight simulation datasets that benefit from such strategies. We consider three different time series-based datasets: \textit{Parking Brake Dataset (PBD)}, \textit{Flap Dataset (FD)} and \textit{Multiple Anomaly Dataset (MAD)} which simulate the application of parking brakes during the takeoff, the deployment of a wrong flap during takeoff and the multiple brake anomalies (\textit{left-brake, right-brake, and auto-brake}) respectively. These datasets are captured on an industry-grade flight simulator. In all these datasets, we train an RNN to predict whether a given sequence is anomalous or not. We compare our method with CA and an approximation to the second-order term in Eqn~\ref{expected taylor expansion eq} proposed in~\citep{ca}. Tab~\ref{tab:total_results} shows the results, highlighting the improvements in time needed to compute ACE in our method. Appendix~\ref{implementation} contains qualitative results showing the performance of these methods using ACE plots.
\begin{table}
    \centering
    \scalebox{0.9}{
    \begin{tabular}{lcccc}
    \toprule
   \textbf{Methods($\downarrow$)/ Datasets($\rightarrow$)} &\textbf{PBD}&\textbf{FD}&\textbf{MAD}\\
    \midrule
        CA (Eqn~\ref{expected taylor expansion eq})&  64.50&69.33&176\\
        CA (Hess.$\times$Cov.) & 20&41&91\\
        $\nabla^2f \approx J^TJ$ &18.5 &24&48.50\\
        \midrule
        Ours + CA  & 21.50&25.5&73\\
        Ours + CA (Hess.$\times$Cov.) & 0.13 &20& 25\\
        Ours + ($\nabla^2f \approx J^TJ$) & 0.004&1&3 \\
    \bottomrule
    \end{tabular}
    }
    \caption{Time taken in minutes by different methods of finding causal effects for flight simulation datasets.}
    \label{tab:total_results}
\end{table}
\section{Conclusions}
\label{sec_conclusions}
We present a new perspective to learn and quantify causal effects in NNs. Using available prior causal knowledge, we design an ante-hoc causal explanation method to study both direct and indirect causal effects of inputs on the output of an NN. The work also presents effective approximation strategies to compute causal effects for high-dimensional data. Our experiments on synthetic and real-world data show significant promise of the methodology to elicit direct and indirect causal effects in an NN model.

\bibliography{aaai24}

\clearpage
\appendix
\setcounter{table}{0}
\renewcommand{\thetable}{A\arabic{table}}

\setcounter{figure}{0}
\renewcommand{\thefigure}{A\arabic{figure}}

\section*{Appendix}

In this Appendix, we include the following additional information, which we could not include in the main paper due to space constraints.
\begin{itemize}
    \item Causality preliminaries are presented in \S~\ref{sec preliminaries}.
    \item Comparison of our method with causal Shapley values is presented in \S~\ref{sec cshap comparison}
    \item Proof of causal effect identifiability is given in \S~\ref{sec identifiability}.
    \item Algorithm for evaluating direct and indirect causal effects in NNs and an algorithm for efficient implementation using the binning approach are outlined in \S~\ref{algorithms}. 
    \item Experimental setup and additional results are in \S~\ref{sec setup and additional results}.
    \item Details on efficient implementation strategies are presented in \S~\ref{implementation}.
    \item Related work on concept-based explainability is discussed in \S~\ref{sec related work concept based}.
    \item A motivating example for learning and explaining indirect causal effects is given in \S~\ref{sec another motivating example}.
\item Uniqueness and usefulness of causal explanations is discussed in \S~\ref{unique_causal}.
\end{itemize}

\section{Causality Preliminaries}
\label{sec preliminaries}
This section provides the basic definitions and concepts required to understand our paper. We start with two popular ways of modeling causality in the real world: (i) structural causal models and (ii) causal graphical models.

\noindent \textbf{Structural Causal Models:} A Structural Causal Model (SCM) $\mathcal{S}(\mathbf{V}, \mathbf{U}, \mathcal{F}, P_{\mathbf{U}})$ encodes cause-effect relationships among a set of random variables $\mathbf{V}\cup \mathbf{U}$ using a set of structural equations $\mathcal{F}$ relating each variable $X\in \mathbf{V}\cup \mathbf{U}$ with its parents $pa(X)\in \mathbf{V}\cup \mathbf{U}\setminus \{X\}$. That is, each variable $X\in \mathbf{V}$ is obtained using the structural equation as $X = f(pa(X))$ for some $f\in \mathcal{F}$. The variables in $\mathbf{U}$ are called exogenous variables denoting uncontrolled external factors. $P_{\mathbf{U}}$ is the probability distribution of exogenous variables. The variables in $\mathbf{V}$ are called endogenous variables. 

\noindent \textbf{Causal Graphical Models:} A causal graphical model $\mathcal{G} (\mathbf{V}\cup \mathbf{U},\mathcal{E})$ consists of a set of vertices $\mathbf{V}\cup \mathbf{U}$ and a set of edges $\mathcal{E}$. The set of edges $\mathcal{E}$ indicate the causal relationships among the variables in $\mathbf{V}\cup \mathbf{U}$. To make an analogy between structural causal models and causal graphical models, the set of vertices $\mathbf{V}$ corresponds to the set of endogenous variables, the set of vertices $\mathbf{U}$ corresponds to the set of exogenous variables, and the set of edges $\mathcal{E}$ corresponds to the set of structural equations $\mathcal{F}$ relating each variable with its parents. Concretely, if $X = f(pa(X));\ f\in\mathcal{F}$, then $\forall X_i \in pa(X)$, we draw a directed edge from $X_i$ to $X$. A \textit{path} in a causal graph is defined as a sequence of unique vertices $X_1, X_2,\dots, X_n$ with an edge between each consecutive vertices $X_i$ and $X_{i+1}$ where the edge between $X_i$ and $X_{i+1}$ can be either $X_i \rightarrow X_{i+1}$ or $X_{i+1}\rightarrow X_i$. A \textit{directed path} is defined as a sequence of unique vertices $X_1,X_2,\dots, X_n$ with an edge between each consecutive vertices $X_i$ and $X_{i+1}$ so that the edge between $X_i$ and $X_{i+1}$ takes the form $X_i \rightarrow X_{i+1}$. If there exists a directed path from $X_i$ to $X_j$, $X_i$ is called an ancestor of $X_j$ and $X_j$ is called a descendant of $X_i$. 

\begin{figure}[H]
    \hspace{-15pt}
    \centering
    \scalebox{0.5}{
    \tikzset{every picture/.style={line width=0.75pt}}  
    \input{images/direct_indirect.tikz}
    }    \caption{Direct and indirect effects of $X_1$ on $X_3$.}
    \label{fig:direct_indirect_effects}
\end{figure}
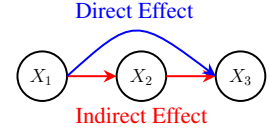

There are three basic causal structures formed with three variables $X_1,X_2,X_3\in \mathbf{V}$: (1)$X_1\rightarrow X_2\rightarrow X_3$, (2) $X_1\leftarrow X_2 \rightarrow X_3$, (3) $X_1\rightarrow X_2\leftarrow X_3$. These three basic causal structures are called 
\textit{chain, fork}, and \textit{collider}, respectively. $X_2$ in $X_1\rightarrow X_2\leftarrow X_3$ is called collider node. $X_1,X_2,X_3$ form a $v$-structure in $\mathcal{G}$ if they form a collider structure $X_1\rightarrow X_2\leftarrow X_3$ with $_1, X_3$ being non-adjacent. A \textit{backdoor} path from $X_i$ to $X_j$ is a path that starts with an arrow into $X_i$ (i.e., a path that starts with $X_i\leftarrow$). A directed path starting from a node $X_i$ and ending at a node $X_j$ is called a \textit{causal path} from $X_i$ to $X_j$. A path that is not a causal path is called a \textit{non-causal path}. For example, the path $X_1\rightarrow X_2\rightarrow X_3$ is a causal path from $X_1$ to $X_3$, and the path $X_1\leftarrow X_2\rightarrow X_3$ is a non-causal path from $X_1$ to $X_3$. 
The causal effect of a variable $X_i$ on another variable $X_j$ that is not flowing through other variables is known as the direct causal effect. The causal effect of a variable $X_i$ on another variable $X_j$ flowing through other variables is known as the indirect causal effect. For example, consider Fig~\ref{fig:direct_indirect_effects}. The causal effect of $X_1$ on $X_3$ that flows through the causal path $X_1\rightarrow X_3$ is the direct causal effect, and the causal effect of $X_1$ on $X_3$ that flows through the causal path $X_1\rightarrow X_2 \rightarrow X_3$ is the indirect causal effect.

\begin{definition}
    \textbf{($d$-Separation.)} A path $p$ between $X_i$ and $X_j$ given $\mathbf{S}\subset \mathbf{V} \setminus \{X_i, X_j\}$ is said to be open, if and only if (1) every collider node on $p$ is in $\mathbf{S}$ or has a descendant in $\mathbf{S}$, and (2) no other non-collider variables on $p$ are in $\mathbf{S}$. If the path $p$ is not open, then $p$ is blocked. $X_i$ and $X_j$ are $d$-separated given $\mathbf{S}$, if and only if every path from $X_i$ to $X_j$ is blocked by $\mathbf{S}$.
\end{definition}

\begin{definition}
\textbf{(The Backdoor Criterion.)} Given a pair of variables $(X_i,X_j)\in \mathbf{V}\times\mathbf{V}$, a set of variables $\mathbf{S}\subset \mathbf{V}$ satisfies the backdoor criterion relative to $(X_i, X_j)$ if no node in $\mathbf{S}$ is a descendant of $X_i$, and $\mathbf{S}$ blocks every backdoor path between $X_i$ and $X_j$.
\end{definition}

\begin{definition} \textbf{(Interventional Distribution~\citep{pearl2009causality}).}
    \label{interventionaldistribution} 
    The interventional distribution of a set of variables $\mathbf{X}=\{X_1,\dots,X_n\}$ under an intervention to $X_i$ with a value $x_i$, denoted by $do(X_i=x_i)$, is defined as follows.
    \begin{equation*}
        p(\mathbf{X}|do(X_i=x_i)) = \begin{cases}
    \prod_{j\neq i}p(X_j|pa(X_j))& \text{if}\ \ X_i= x_i\\
    0              & \text{if}\ \ X_i\neq x_i
            \end{cases}
    \end{equation*}
\end{definition}
The resulting probability distribution of a set of variables $\mathbf{X}_{\setminus i}=\{X_1,\dots,X_n\}\setminus \{X_i\}$ under the intervention $do(X_i=x_i)$ is same as the probability distribution of $\mathbf{X}_{\setminus i}$ induced by the \textit{intervened/manipulated} causal graph $\mathcal{G}_{do(X_i)}$.  $\mathcal{G}_{do(X_i)}$ is obtained by removing all incoming arrows to $X_i$ in $\mathcal{G}$~\citep{pearl2009causality}.

\begin{definition}
\label{def:validadjustment}
    \textbf{(Valid Adjustment Set.)}
    A set of variables $\mathbf{S}$ is said to be a valid adjustment set for calculating $ACE_X^Y$ if and only if $p(Y|do(X=x)) = \mathbb{E}_{\mathbf{s}\sim\mathbf{S}} [p(Y|X=x, \mathbf{S}=\mathbf{s})]$
\end{definition}
To find the $ACE^Y_X$ w.r.t. two interventions $x,x^*$, we perform the backdoor adjustment using a valid adjustment set $\mathbf{S}$ as follows.
\begin{equation*}
\begin{aligned}
    ACE^Y_X & \coloneqq \mathbb{E}[Y|do(X=x)]-\mathbb{E}[Y|do(X=x^*)]\\
    &= \mathbb{E}_{\mathbf{S}}\mathbb{E}[Y|X=x, \mathbf{S}]-\mathbb{E}_{\mathbf{S}}\mathbb{E}[Y|X=x^*,\mathbf{S}]\\
\end{aligned}
\end{equation*}
\begin{table*}
    \centering
    \scalebox{0.93}{
    \begin{tabular}{lccllc}
    \toprule
        \textbf{Dataset} & \textbf{Num. of Features} & \textbf{Num. Data Points} & \textbf{Architecture} & \textbf{Application} & \textbf{Algorithms Studied} \\
    \midrule
         Synthetic data& 3 & 1000 & Feedforward neural network& Regression& ~\ref{algo:nn_edges_input_layer},~\ref{algo:ace}\\
         AutoMPG& 5 & 392 & Feedforward neural network& Regression& ~\ref{algo:nn_edges_input_layer},~\ref{algo:ace}\\
         Lung Cancer & 7 & 10,000 & Feedforward neural network& Classification& ~\ref{algo:nn_edges_input_layer},~\ref{algo:ace}\\
         SACHS& 11 & 10,000 & Feedforward neural network&Classification & ~\ref{algo:nn_edges_input_layer},~\ref{algo:ace}\\
         Parking brake& 48 & 3360 & Recurrent neural network&Classification&\ref{algo:improvements}\\
         Flap& 96 & 6370 & Recurrent neural network&Classification&\ref{algo:improvements}\\
         Multiple anomaly& 52 & 4403 & Recurrent neural network&Classification&\ref{algo:improvements}\\
    \bottomrule
    \end{tabular}
    }
    \caption{Summary of datasets and applications}
    \label{tab:datasets}
\end{table*}
\section{Comparison with causal Shapley values}
\label{sec cshap comparison}
In this section, we show that the causal Shapley value~\citep{causalshapley} of a feature $X_i$ on the output $\hat{Y}$ of an NN differs from the causal effect of $X_i$ on the output. Consider the following formula that computes the causal Shapley value $\phi_i$ of a feature $X_i$.
\begin{equation}
\label{shapleyformula}
\small{
    \phi_i = \displaystyle \sum_{S\subseteq \{1..N\}\setminus\{i\}} \frac{|S|!(N-|S|-1)!}{N!}\left[ v(S\cup \{i\})-v(S) \right]
}
\end{equation}
Where $N$ is the number of features and $v(S)$ is defined as $v(S)=\mathbb{E}[\hat{Y}|do(\mathbf{X}_S=\mathbf{x}_s)]$. Now consider the expression for the causal effect of $X_i$ on the output $\hat{Y}$.
\begin{equation}
\label{aceformula}
    ACE_{X_i}^Y = \mathbb{E}[\hat{Y}|do(X_i=x_i)]-\mathbb{E}[\hat{Y}|do(X_i=x_i^*)]
\end{equation}
In Eqn~\ref{shapleyformula}, intervention is performed on the set of \textit{in-coalition} features $S$ and the feature $X_i$, irrespective of their causal relationship with the feature $X_i$. However, in the expression for average causal effect (Eqn~\ref{aceformula}), intervention is performed on only the feature $X_i$, leading to a different value obtained from the Shapley formula.

\section{Causal effect identifiability in $\mathcal{N}$ and $\mathcal{N}^{Ind}$}
\label{sec identifiability}
This section provides formal proof of the identifiability of causal effects in neural networks. We first formally prove why the proposed Hypothesis~\ref{hypothesis1} is required to study indirect causal effects in an NN. Consider an NN $\mathcal{N}$ with input features $X_1,\dots,X_n$ and an output neuron $\hat{Y}$ in the final layer.  We can marginalize the hidden layers of $\mathcal{N}$ and consider $\hat{Y}$ as a function $f$ of only $\mathcal{N}$'s inputs i.e., $\hat{Y}=f(X_1,\dots,X_n)$. In a simple feed-forward NN $\mathcal{N}$, the inputs are not causally connected by design i.e., no NN edges among input neurons. If we perform an intervention $do(X_i=x_i)$ in $\mathcal{N}$, it will not affect any other input features but affects $\hat{Y}$ through the function $f$. That is, $X_i$ has only a direct effect on $\hat{Y}$ (Defn~\ref{def:direct_effect}), and indirect effect of $X_i$ on $\hat{Y}$ does not exist. If we introduce NN edges among inputs, the intervention $do(X_i=x_i)$ will influence $X_i$'s descendants via the newly introduced edges in layer $0$ and influence $\hat{Y}$ via the function $f$. Hence the indirect causal effect of $X_i$ on $\hat{Y}$ can be identified, as shown in the next paragraph. Now, to \textit{quantify} the indirect causal effects, we need to rely on the effect of $X_i$ on $\hat{Y}$ via the children $ch(X_i)$ of $X_i$. For this purpose, it is required to learn the weights corresponding to the edges between $X_i$ and $ch(X_i)$ while optimizing $\mathcal{N}$'s objective, such as minimizing regression loss.
\begin{algorithm}
\footnotesize
\caption{\small Pseudocode for computing $\mathbb{E}[\hat{Y}|do(X_i^*, \mathbf{Z}_{X_i^*})]$, $\mathbb{E}[\hat{Y}|do(X_i, \mathbf{Z}_{X_i^*})]$, $\mathbb{E}[\hat{Y}|do(X_i^*, \mathbf{Z}_{X_i})]$ for $\mathcal{N}^{Ind}$ model}
   \label{algo:ace}
\begin{algorithmic}[1]

   \STATE {\bfseries Input:} NN output $\hat{Y}$, $X_i$,  $Z=ch(X_i)$, interventions range [$l, h$], number of interventions $n$, $f_0^p$ learned in $l_0$ for each $X_p$, baseline intervention $x_i^*$. 
    \STATE {\bfseries Output:} {\footnotesize  $\mathbb{E}[\hat{Y}|do(X_i^*, \mathbf{Z}_{X_i^*})]$, $\mathbb{E}[\hat{Y}|do(X_i, \mathbf{Z}_{X_i^*})]$, $\mathbb{E}[\hat{Y}|do(X_i^*, \mathbf{Z}_{X_i})]$}
   \STATE {\bfseries Initialize:} $Cov[X_i,:] = 0$, $Cov[:,X_i] = 0$, $IE = []$, $x_i = low^i; \mu = [\mathbb{E}(X_1), \dots, \mathbb{E}(X_n)]$

   \WHILE{$x_i \leq h$}
    \STATE {\textbf{case 1:} To compute $\mathbb{E}[\hat{Y}|do(X_i^*, \mathbf{Z}_{X_i^*})]$} 
    \STATE{\hspace{1cm} $X_i = x_i^*$ \hspace{1cm} \textbackslash * Defns~\ref{def:direct_effect} and~\ref{def:indirect_effects} * \textbackslash}
    \STATE{\hspace{1cm}  $\mu[i] = x_i^*$}

    \STATE {\textbf{case 2:} To compute $\mathbb{E}[\hat{Y}|do(X_i, \mathbf{Z}_{X_i^*})]$}
    \STATE{\hspace{1cm} $X_i = x_i^*$ \hspace{1cm} \textbackslash * Defn~\ref{def:direct_effect} * \textbackslash}
    \STATE{\hspace{1cm}  $\mu[i] = x_i$}

    \STATE {\textbf{case 3:} To compute $\mathbb{E}[\hat{Y}|do(X_i^*, \mathbf{Z}_{X_i})]$}
    \STATE{\hspace{1cm} $X_i = x_i$ \hspace{1cm} \textbackslash * Defn~\ref{def:indirect_effects} * \textbackslash }
    \STATE{\hspace{1cm}  $\mu[i] = x_i^*$}
    
    \FOR{ $X_p\in \mathbf{Z}$}
       \STATE{$X_p = f_0^p(pa(X_p))$}
        \STATE{$\mu[p] = \mathbb{E}_{pa(X_p)}\big[\mathbb{E}[X_p|X_i,pa(X_p)\setminus \{X_i\}]\big]$}
        \STATE{Compute $Cov[X_j,X_p]\ \forall j\neq i,p$}
    \ENDFOR
    
       \STATE append $f(\mu)$+ $\frac{1}{2}$Trace($\nabla^2f(\mu)\times Cov$) to $IE$\;
       \STATE $x_i := x_i + \frac{h - l}{n}$ \;
   \ENDWHILE
 \STATE{return $IE$}
\end{algorithmic}
\end{algorithm}
We now show that the graphical criterion to experimentally identify the causal effects are satisfied in $\mathcal{N}^{Ind}$, which is enough to claim that causal effects are identifiable in $\mathcal{N}^{Ind}$~\citep{pearl2009causality}. Let $\mathcal{I}_v = \{X_i, Z, \hat{Y}, W\}$ be the set of nodes in the marginalized NN where $Z$ is the set of nodes in the input layer that are in a path from $X_i$ to $\hat{Y}$, and $W$ is the set of remaining nodes in the input layer. We need to show that $\big[\hat{Y}|do(X_i, Z)\big]\perp \!\!\!\perp Z_{X_i^*}|W$ for direct causal effect identifiability and $\big[\hat{Y}|do(X_i^*, Z)\big]\perp \!\!\!\perp Z_{X_i}|W$ for indirect causal effect identifiability in $\mathcal{N}^{Ind}$ model~\citep{pearl2001direct,pearl2009causality}. Graphically, proving these two criteria is equivalent to showing $(\hat{Y}\perp \!\!\!\perp Z|W)_{\mathcal{G}_{\underline{X_i Z}}}$~\citep{pearl2001direct,pearl2009causality}. Here $\mathcal{G}_{\underline{X_i Z}}$ is formed by removing from $\mathcal{G}$ the outgoing edges from $X_i$ and from the nodes of $Z$. In $\mathcal{N}^{Ind}$ model, after removing the outgoing edges from $X_i$ and from the nodes of $Z$, there will be no directed edges from $X_i, Z$ to $\hat{Y}$ and all the \textit{backdoor} paths to $\hat{Y}$ (if exists) will be blocked by $W$ and hence satisfy the property $(\hat{Y}\perp \!\!\!\perp Z|W)_{\mathcal{G}_{\underline{X_i Z}}}$. On the other hand, since there is no indirect causal paths in the SCM of $\mathcal{N}$, indirect causal effects do not exist in $\mathcal{N}$ i.e., indirect causal effects in $\mathcal{N}$ are $0$.

\section{Algorithms for Causal Effect Evaluation and Efficient Implementation Using Binning}
\label{algorithms}

Algorithm~\ref{algo:ace} outlines the steps to evaluate $\mathbb{E}[\hat{Y}|do(X_i^*, \mathbf{Z}_{X_i^*})]$, $\mathbb{E}[\hat{Y}|do(X_i, \mathbf{Z}_{X_i^*})]$, $\mathbb{E}[\hat{Y}|do(X_i^*, \mathbf{Z}_{X_i})]$ for $\mathcal{N}^{Ind}$. Algorithm~\ref{algo:improvements} outlines the binning methodology to improve the run time requirements to find causal effects.
\begin{algorithm}
\footnotesize
\caption{Pseudocode for efficient evaluation of $ACE_{X_i}^{\hat{Y}}$}
\label{algo:improvements}
\begin{algorithmic}[1]
\STATE {\bfseries Input:} Test sample $\mathbf{X}_{te}$, $X_i$, Intervention value $\alpha$, Database $DB$ to store offline interventional statistics, Training data $\mathcal{D}$, Nearest neighbor function $NB$.
\STATE {\bfseries Output:} $\mathbb{E}[\hat{Y}|do(X_i=\alpha)]$
\STATE{\textbackslash * Offline Phase * \textbackslash}
\FOR{$\mathbf{X}_{tr} \in \mathcal{D}$}
\STATE {$DB[\mathbf{X}_{tr};E;x_i]=\mathbb{E}[X_j|do(X_i)]\ \forall j, x_i$ }
\STATE {$DB[\mathbf{X}_{tr};C;x_i]=Covariance(X_j, X_l|do(X_i))\ \forall j,l,x_i$}
\ENDFOR

\STATE{\textbackslash * Online Phase * \textbackslash}
\STATE {\textbf{step 1:} $\mathbf{X}_{tr}=NB(\mathbf{X}_{te}, \mathcal{D})$}

\STATE {\textbf{step 2:} $x_i = NB(\beta)$}

\STATE {\textbf{step 3:} $\mathbb{E}[X_j|do(X_i)]\ \forall j = DB[\mathbf{X}_{tr};E;x_i]$
}

\STATE {\textbf{step 4:} $Cov(X_j, X_l|do(X_i))\ \forall j,l = DB[\mathbf{X}_{tr};C;x_i]$}

\STATE {\textbf{step 5:} Evaluate $ACE_{X_i}^{\hat{Y}}$ using $\mathbb{E}[X_j|do(X_i)]\ \forall j$, $Covariance(X_j,X_l|do(X_i)))\ \forall j,l$}

\STATE {return $ACE_{X_i}^{\hat{Y}}$}
\end{algorithmic}
\end{algorithm}

\section{Experimental Setup and Additional Results}
\label{sec setup and additional results}
In this section, we present details on the experimental setup. Tab~\ref{tab:datasets} shows the details on the datasets and applications we study in this paper. In the Sachs dataset, we compute causal effects of inputs on the first output neuron among three neurons in the final layer. Integrated Gradients (IG), Causal Attributions (CA), and Causal Shapley (CSHAP) explanations are obtained on a model trained with empirical risk minimization loss (ERM). Apart from ERM, we train CREDO, a causal regularization method, and our proposed AHCE method. From the results in the main paper, the proposed AHCE method gives better causal explanations than baselines with no change in downstream performance. Tab~\ref{tab:performance} supports the fact that there is no significant difference in the performance of the three models: ERM, CREDO, and AHCE. If an input feature is real-valued, we consider 1000 interventions (i.e., 1000 $X_i$ values in Eqn~\ref{expected taylor expansion eq}) in its input range to calculate interventional expectation values. Across all our experiments, we use Adam optimizer with default parameter values except for weight decay, which we set to $0.0001$. Batch sizes for the Synthetic, Auto-MPG, Lung Cancer, and Sachs experiments are 1, 1, 16, and 64, respectively. We get the best results with 20 training epochs for experiments performed on Synthetic and AutoMPG datasets, whereas we observe that 50 epochs are required in the case of SACHS and Lung Cancer to obtain good results. For the Integrated Gradients method (IG), we set the \textit{n\textunderscore steps} parameter to be 50 across all experiments. Since obtaining Shapley values is computationally expensive, we pick 100 values randomly from the dataset to compute them and subsequently calculate the performance metrics (we observe that the number of data points has a minor impact on the overall Shapley values).

\begin{table}
    \centering
    \footnotesize
    \scalebox{0.80}{
    \begin{tabular}{l|cccc}
    \textbf{Dataset($\rightarrow$)}&\textbf{Synthetic}&\textbf{Auto-MPG}&\textbf{Lung Cancer}&\textbf{Sachs}\\
       \textbf{Method($\downarrow$)}& (RMSE ($\downarrow$))&(RMSE ($\downarrow$))& (Accuracy ($\uparrow$))&(Accuracy ($\uparrow$))\\
       \midrule
        ERM&0.008$\pm$0.00&0.008$\pm$0.00&85.09$\pm$0.00&81.55$\pm$0.18\\
        CREDO &0.012$\pm$0.00&0.016$\pm$0.00&84.09$\pm$0.03&81.69$\pm$0.00\\
        AHCE&0.010$\pm$0.00&0.013$\pm$0.00&85.09$\pm$0.00&81.45$\pm$0.02\\
        \bottomrule
    \end{tabular}
    }
    \caption{Performance of methods on various datasets.}
    \label{tab:performance}
\end{table}
\begin{table}
\footnotesize
    \centering
\scalebox{0.95}{
    \begin{tabular}{cc|cc}
         &\textbf{Feature}& \textbf{CSHAP}~\citepalias{causalshapley} & \textbf{AHCE} (Ours)\\
         \midrule
         &W&3.26$\pm$0.00&3.09$\pm$0.00\\
         &Z&1.82$\pm$0.00&1.89$\pm$0.00\\
         &X&0.09$\pm$0.00&0.02$\pm$0.00\\
         \cmidrule{2-4}
         \parbox[t]{2mm}{\multirow{-4}{*}{\rotatebox[origin=c]{90}{RMSE ($\downarrow$)}}}&Average&1.72$\pm$0.00&\cellcolor{gray!50}\textbf{1.67$\pm$0.01}\\
         \midrule
         &W&4.52$\pm$0.00&4.74$\pm$0.05\\
         &Z&3.71$\pm$0.00&3.81$\pm$0.00\\
         &X&0.16$\pm$0.00&0.04$\pm$0.04\\
         \cmidrule{2-4}
         \parbox[t]{2mm}{\multirow{-4}{*}{\rotatebox[origin=c]{90}{Frechet ($\downarrow$)}}}&Average&\cellcolor{gray!50}\textbf{2.80$\pm$0.00}&2.86$\pm$0.03\\
         \bottomrule
    \end{tabular}
    }
    \caption{Comparison of CSHAP and AHCE w.r.t. indirect causal effects on synthetic dataset.}
    \label{tab:indirect comparison}
\end{table}

\begin{figure}
    \centering
    \includegraphics[width=0.5\textwidth]{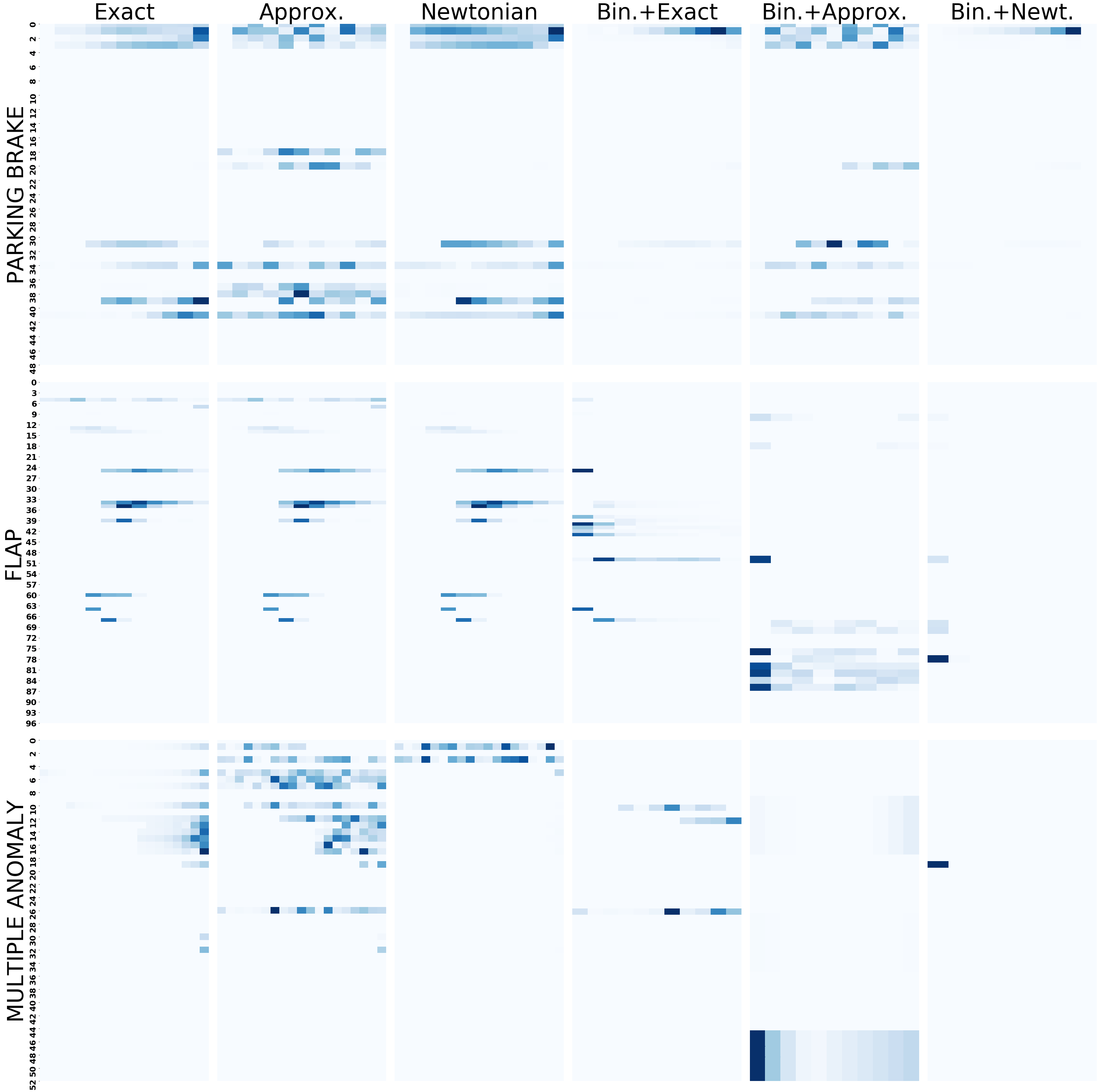}
    \caption{Results on binning approximation to causal explanations for time series data}
    \label{fig:binning_results}
\end{figure}

The main paper contains results w.r.t. total causal effects. We present the results w.r.t. indirect causal effects in Tab~\ref{tab:indirect comparison}. As noted in the main paper \S~\ref{sec experiments}, the indirect effects do not exist for IG, CA, and CREDO. For real-world datasets, we do not have access to the underlying structural equations; hence, we cannot evaluate ground truth indirect causal effects. Figure~\ref{fig:binning_results} shows the absolute ACE values of each feature at different time steps for three time-series datasets and baseline methods. In Fig~\ref{fig:binning_results}, `Exact` method refers to the evaluation of Eqn~\ref{expected taylor expansion eq}, `Approx` method refers to the approximation to the second order term in Eqn~\ref{expected taylor expansion eq} proposed in~\citep{ca}, `Newtonian` method refers to the Newtonian approximation $\nabla^2f \approx J^TJ$. For each plot in Fig~\ref{fig:binning_results}, x-axis denotes the time step (unrolling length of RNN), and y-axis denotes the features. It can be seen that, in many cases, the proposed approximation's causal explanations (last four columns) are close to the exact computation of causal explanations (first two columns). However, there is a trade-off between the binning parameters (number of bins, distance function used in clustering methods, etc.) and the obtained causal explanations.

\section{Efficient Implementation Strategies}
\label{implementation}
\begin{figure}
    \centering
    \includegraphics[width=0.5\textwidth]{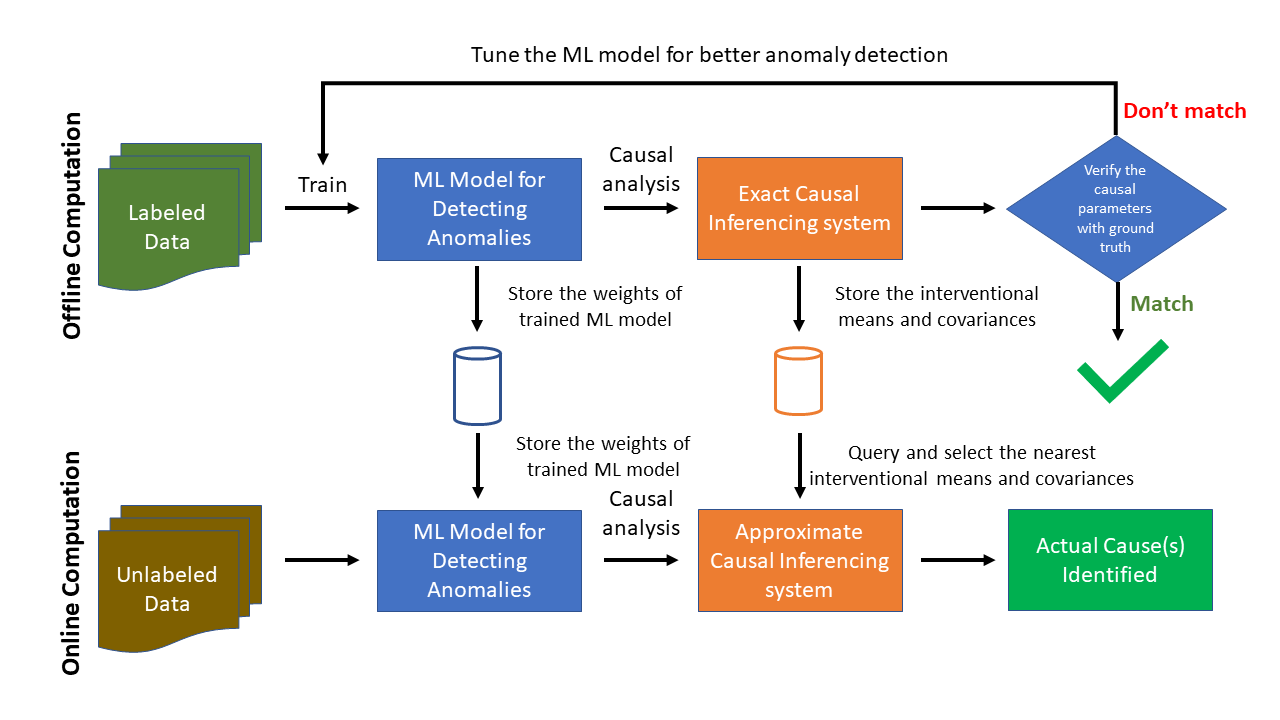}
    \caption{Binnig (offline - online) framework for quick and accurate explainability}
    \label{unifiedframework}
\end{figure}

In this section, we expound on the strategies to reduce the storage requirements for binning approach (Sec~\ref{sec:improve_time_and_space_complexity} of main paper). The overall approach of online-offline (binning) method is outlined in Figure~\ref{unifiedframework}. Storing offline interventional statistics for every point on the dataset becomes impractical.
This is owing to the multiple aspects of data at each time step of computing interventional expectation: the number of interventional values, the number of features, and in the case of RNNs, the unrolling length of RNN. Even if storage bottlenecks are not a concern, managing a data structure that can address, log, and access this on-demand becomes tedious and counter-intuitive. So we seek to reduce the number of points, the number of interventions for each point, and the amount of data stored per point while maintaining the accuracy of online results of incoming points.

We get data clusters so that the cluster centers can be used as a proxy for cluster points so that the number of points to compare with a test data point during the online phase are significantly less. However, the success of clustering methods differs greatly owing to many factors. One important consideration is the distance function used. Euclidean distances are standard and, owing to current experimental hurdles, are used herein experiments. But ideally, a distance metric that depends on the accuracy of the ACE values (entire pipeline from offline to online) is ideal. We seek to find clustering methods on the dataset with a minimum accuracy trade-off while maximizing the number of points clustered, thus reducing storage by the average size of the cluster. We use the following clustering methods in this work.

\noindent \textbf{KD Tree:}
A K-Dimensional (KD) tree is constructed with flattened points so that cluster centroids can be extracted and used for offline computation. Here a hyperparameter is the maximum distance between two points for them to be considered neighbors. Our experiments suggest a Euclidean distance of 10 is ideal, but a roll-off till 20 is still acceptable.

\noindent \textbf{DBSCAN:}
Since the nature and shape of the clusters are unknown, we can use DBSCAN to get a density-based estimate, where points too sparsely connected to an existing cluster would be considered as outliers. This adds another hyperparameter apart from the maximum distance, called minimum points, that quantifies the number of nearby points needed to consider it as a cluster. This allows us to have a handle on the storage requirements but setting this value too high results in many points being outliers without a cluster.

\noindent \textbf{Complexity Analysis in RNNs}
In RNNs, the number of steps to unroll the network, call the \textit{chunk} size, impacts the dataset size when we collect data from a stream. From a data stream, we split the dataset such that chunks are overlapping sub-sequences i.e., if the length of a stream is $l$ and chunk size is $k$, we get approximately $l-k$ chunks in total. Evaluating Eqn~\ref{interventional eq} takes order of $\mathcal{O}(l\times n^p)$ time (assuming $k<<l$ and features are $p$-valued, $n$ is the of input size). On the other hand, evaluating the approximation in Eqn~\ref{expected taylor expansion eq} also scales in the order of $\mathcal{O}(kn)$ as the inputs at a particular time step affect the inputs at the next time step in RNNs. In RNNs, interventional statistics calculation takes considerable time as these steps have to be done for every feature in the data point at every time step starting from $t$ to 1, where $t$ is the time step at which we wish to calculate the causal effect, which would take considerable time.

\section{Related Work on Concept-Based Explanations}
\label{sec related work concept based}
Concept-based explanation models attempt to explain an NN model behavior in terms of semantically meaningful concepts in the input when the input is not readily expressed as a discrete set of concepts. Concept bottleneck models~\citep{cbmodels}, and concept embedding models~\citep{cemodels} learn the relationship between manually annotated concepts in the input samples and the model predictions. Linguistic concept-based explanation models, such as~\citep{flexmodel,ccnngc} try to automatically discover semantically meaningful concepts in the input space by leveraging text annotations corresponding to the inputs and using them to explain model predictions. Quantitative testing with concept activation vectors (TCAV)~\citep{tcav} can be used as a post hoc method to explain model predictions in terms of arbitrary concepts that may exist in the input samples but are not known at the time of training.~\citep{eecv18} is a method to decompose the input samples into semantically interpretable concepts derived from a pre-trained concept corpus. However, as discussed earlier, when the number of concepts is more, evaluating the causal explanations is time-consuming. In this work, we tackle this time complexity issue using the binning approach as explained in Sec~\ref{sec:improve_time_and_space_complexity} and in Appendix \S~\ref{implementation}.

\section{Motivating Example for Modeling Indirect Causal Effecs}
\label{sec another motivating example}
Besides the illustrative example presented in section \S~\ref{sec introduction} of the main paper, this section introduces another motivating example that underscores the importance of contemplating indirect causal effects within an NN model.

Consider predicting the health condition $H$ of a patient using their age $A$ and consumed medicine dose $M$. In the real-world, the age group may determine the medicine dosage value to administer, i.e., $A\rightarrow M$ (e.g., adults are administered higher dosages as compared to children). Also, $A$ and $M$ have direct causal effects on $H$. That is $A\rightarrow H, M\rightarrow H$. However, in a simple feed-forward NN, we do not have $A\rightarrow M$. If we train an ante-hoc model by adding the edge $A\rightarrow M$ to the NN, we can study the indirect effects of $A$ on $H$. In this case, it is desirable for $A$ to have non-zero direct and indirect causal effects on the output. If any one of the direct or indirect causal effects of $A$ on $H$ is zero or very small in magnitude, we can conclude that the model has not learned correct relationships. We can then use our explanations to study data biases or to debug the model for potential improvement of model performance while maintaining real-world causal relationships in the NN model.

\section{Uniqueness and Usefulness of Causal Explanations}
\label{unique_causal}
In this section, we discuss how causal explanations are different and much more useful when compared to any other explanations in safety-critical applications. Consider the example of a clinical decision support application. When an NN model is used to predict the \textit{risk of cardiac disease}, a high importance for an input variable such as \textit{hypertension} may arise due to its high correlation with \textit{risk of cardiac disease}. However, the person's \textit{diet choice} (another input variable) may be causing both \textit{hypertension} and \textit{risk of cardiac disease}. Modeling and studying causal effects in NN models help a medical practitioner take appropriate decisions, compared to the reliance on non-causal explanations.

\end{document}

%% file: images/introduction.tikz
\begin{tikzpicture}[x=0.75pt,y=0.75pt,yscale=-1,xscale=1]

\draw  [line width=1.5]  (30,125) .. controls (30,111.19) and (41.19,100) .. (55,100) .. controls (68.81,100) and (80,111.19) .. (80,125) .. controls (80,138.81) and (68.81,150) .. (55,150) .. controls (41.19,150) and (30,138.81) .. (30,125) -- cycle ;
\draw  [line width=1.5]  (210,125) .. controls (210,111.19) and (221.19,100) .. (235,100) .. controls (248.81,100) and (260,111.19) .. (260,125) .. controls (260,138.81) and (248.81,150) .. (235,150) .. controls (221.19,150) and (210,138.81) .. (210,125) -- cycle ;
\draw  [line width=1.5]  (120,125) .. controls (120,111.19) and (131.19,100) .. (145,100) .. controls (158.81,100) and (170,111.19) .. (170,125) .. controls (170,138.81) and (158.81,150) .. (145,150) .. controls (131.19,150) and (120,138.81) .. (120,125) -- cycle ;
\draw  [line width=1.5]  (120,222) .. controls (120,208.19) and (131.19,197) .. (145,197) .. controls (158.81,197) and (170,208.19) .. (170,222) .. controls (170,235.81) and (158.81,247) .. (145,247) .. controls (131.19,247) and (120,235.81) .. (120,222) -- cycle ;
\draw [line width=1.5]    (55,150) -- (141.45,195.15) ;
\draw [shift={(145,197)}, rotate = 207.57] [fill={rgb, 255:red, 0; green, 0; blue, 0 }  ][line width=0.08]  [draw opacity=0] (13.4,-6.43) -- (0,0) -- (13.4,6.44) -- (8.9,0) -- cycle    ;
\draw [line width=1.5]    (145,150) -- (145,193) ;
\draw [shift={(145,197)}, rotate = 270] [fill={rgb, 255:red, 0; green, 0; blue, 0 }  ][line width=0.08]  [draw opacity=0] (13.4,-6.43) -- (0,0) -- (13.4,6.44) -- (8.9,0) -- cycle    ;
\draw [line width=1.5]    (235,150) -- (148.55,195.15) ;
\draw [shift={(145,197)}, rotate = 332.43] [fill={rgb, 255:red, 0; green, 0; blue, 0 }  ][line width=0.08]  [draw opacity=0] (13.4,-6.43) -- (0,0) -- (13.4,6.44) -- (8.9,0) -- cycle    ;
\draw  [line width=1.5]  (300,125) .. controls (300,111.19) and (311.19,100) .. (325,100) .. controls (338.81,100) and (350,111.19) .. (350,125) .. controls (350,138.81) and (338.81,150) .. (325,150) .. controls (311.19,150) and (300,138.81) .. (300,125) -- cycle ;
\draw  [line width=1.5]  (480,125) .. controls (480,111.19) and (491.19,100) .. (505,100) .. controls (518.81,100) and (530,111.19) .. (530,125) .. controls (530,138.81) and (518.81,150) .. (505,150) .. controls (491.19,150) and (480,138.81) .. (480,125) -- cycle ;
\draw  [line width=1.5]  (390,125) .. controls (390,111.19) and (401.19,100) .. (415,100) .. controls (428.81,100) and (440,111.19) .. (440,125) .. controls (440,138.81) and (428.81,150) .. (415,150) .. controls (401.19,150) and (390,138.81) .. (390,125) -- cycle ;
\draw  [line width=1.5]  (390,222) .. controls (390,208.19) and (401.19,197) .. (415,197) .. controls (428.81,197) and (440,208.19) .. (440,222) .. controls (440,235.81) and (428.81,247) .. (415,247) .. controls (401.19,247) and (390,235.81) .. (390,222) -- cycle ;
\draw [line width=1.5]    (325,150) -- (411.45,195.15) ;
\draw [shift={(415,197)}, rotate = 207.57] [fill={rgb, 255:red, 0; green, 0; blue, 0 }  ][line width=0.08]  [draw opacity=0] (13.4,-6.43) -- (0,0) -- (13.4,6.44) -- (8.9,0) -- cycle    ;
\draw [line width=1.5]    (415,150) -- (415,193) ;
\draw [shift={(415,197)}, rotate = 270] [fill={rgb, 255:red, 0; green, 0; blue, 0 }  ][line width=0.08]  [draw opacity=0] (13.4,-6.43) -- (0,0) -- (13.4,6.44) -- (8.9,0) -- cycle    ;
\draw [line width=1.5]    (505,150) -- (418.55,195.15) ;
\draw [shift={(415,197)}, rotate = 332.43] [fill={rgb, 255:red, 0; green, 0; blue, 0 }  ][line width=0.08]  [draw opacity=0] (13.4,-6.43) -- (0,0) -- (13.4,6.44) -- (8.9,0) -- cycle    ;
\draw [line width=1.5]    (325,150) .. controls (378.58,166.92) and (392.4,159.1) .. (411.31,151.46) ;
\draw [shift={(415,150)}, rotate = 159.15] [fill={rgb, 255:red, 0; green, 0; blue, 0 }  ][line width=0.08]  [draw opacity=0] (13.4,-6.43) -- (0,0) -- (13.4,6.44) -- (8.9,0) -- cycle    ;
\draw [line width=1.5]    (415,150) .. controls (468.58,166.92) and (482.4,159.1) .. (501.31,151.46) ;
\draw [shift={(505,150)}, rotate = 159.15] [fill={rgb, 255:red, 0; green, 0; blue, 0 }  ][line width=0.08]  [draw opacity=0] (13.4,-6.43) -- (0,0) -- (13.4,6.44) -- (8.9,0) -- cycle    ;
\draw [line width=1.5]    (325,150) .. controls (364.6,168.32) and (423.8,190.06) .. (502.61,151.2) ;
\draw [shift={(505,150)}, rotate = 153.15] [fill={rgb, 255:red, 0; green, 0; blue, 0 }  ][line width=0.08]  [draw opacity=0] (13.4,-6.43) -- (0,0) -- (13.4,6.44) -- (8.9,0) -- cycle    ;

\draw (45.5,117.9) node [anchor=north west][inner sep=0.75pt]  [font=\huge]  {$S$};
\draw (132.5,116.9) node [anchor=north west][inner sep=0.75pt]  [font=\huge]  {$E$};
\draw (224,117.9) node [anchor=north west][inner sep=0.75pt]  [font=\huge]  {$R$};
\draw (138.17,213.4) node [anchor=north west][inner sep=0.75pt]  [font=\huge]  {$I$};
\draw (134,257) node [anchor=north west][inner sep=0.75pt]  [font=\LARGE] [align=left] {(a)};
\draw (315.5,117.9) node [anchor=north west][inner sep=0.75pt]  [font=\huge]  {$S$};
\draw (402.5,116.9) node [anchor=north west][inner sep=0.75pt]  [font=\huge]  {$E$};
\draw (493.67,117.9) node [anchor=north west][inner sep=0.75pt]  [font=\huge]  {$R$};
\draw (408.17,213.4) node [anchor=north west][inner sep=0.75pt]  [font=\huge]  {$I$};
\draw (406,257) node [anchor=north west][inner sep=0.75pt]  [font=\LARGE] [align=left] {(b)};

\end{tikzpicture}

%% file: images/ahce.tikz
\begin{tikzpicture}[x=0.75pt,y=0.75pt,yscale=-1,xscale=1]

\draw  [line width=1.5]  (5,147.25) .. controls (5,134.96) and (14.85,125) .. (27,125) .. controls (39.15,125) and (49,134.96) .. (49,147.25) .. controls (49,159.54) and (39.15,169.5) .. (27,169.5) .. controls (14.85,169.5) and (5,159.54) .. (5,147.25) -- cycle ;
\draw  [line width=1.5]  (163.4,147.25) .. controls (163.4,134.96) and (173.25,125) .. (185.4,125) .. controls (197.55,125) and (207.4,134.96) .. (207.4,147.25) .. controls (207.4,159.54) and (197.55,169.5) .. (185.4,169.5) .. controls (173.25,169.5) and (163.4,159.54) .. (163.4,147.25) -- cycle ;
\draw  [line width=1.5]  (84.2,147.25) .. controls (84.2,134.96) and (94.05,125) .. (106.2,125) .. controls (118.35,125) and (128.2,134.96) .. (128.2,147.25) .. controls (128.2,159.54) and (118.35,169.5) .. (106.2,169.5) .. controls (94.05,169.5) and (84.2,159.54) .. (84.2,147.25) -- cycle ;
\draw  [line width=1.5]  (84.2,233.59) .. controls (84.2,221.3) and (94.05,211.34) .. (106.2,211.34) .. controls (118.35,211.34) and (128.2,221.3) .. (128.2,233.59) .. controls (128.2,245.88) and (118.35,255.84) .. (106.2,255.84) .. controls (94.05,255.84) and (84.2,245.88) .. (84.2,233.59) -- cycle ;
\draw [line width=1.5]    (106.2,169.5) -- (106.2,207.34) ;
\draw [shift={(106.2,211.34)}, rotate = 270] [fill={rgb, 255:red, 0; green, 0; blue, 0 }  ][line width=0.08]  [draw opacity=0] (13.4,-6.43) -- (0,0) -- (13.4,6.44) -- (8.9,0) -- cycle    ;
\draw [line width=1.5]    (185.4,169.5) -- (109.74,209.47) ;
\draw [shift={(106.2,211.34)}, rotate = 332.16] [fill={rgb, 255:red, 0; green, 0; blue, 0 }  ][line width=0.08]  [draw opacity=0] (13.4,-6.43) -- (0,0) -- (13.4,6.44) -- (8.9,0) -- cycle    ;
\draw  [line width=1.5]  (235,147.25) .. controls (235,134.96) and (244.85,125) .. (257,125) .. controls (269.15,125) and (279,134.96) .. (279,147.25) .. controls (279,159.54) and (269.15,169.5) .. (257,169.5) .. controls (244.85,169.5) and (235,159.54) .. (235,147.25) -- cycle ;
\draw  [line width=1.5]  (393.4,147.25) .. controls (393.4,134.96) and (403.25,125) .. (415.4,125) .. controls (427.55,125) and (437.4,134.96) .. (437.4,147.25) .. controls (437.4,159.54) and (427.55,169.5) .. (415.4,169.5) .. controls (403.25,169.5) and (393.4,159.54) .. (393.4,147.25) -- cycle ;
\draw  [line width=1.5]  (314.2,147.25) .. controls (314.2,134.96) and (324.05,125) .. (336.2,125) .. controls (348.35,125) and (358.2,134.96) .. (358.2,147.25) .. controls (358.2,159.54) and (348.35,169.5) .. (336.2,169.5) .. controls (324.05,169.5) and (314.2,159.54) .. (314.2,147.25) -- cycle ;
\draw  [line width=1.5]  (314.2,233.59) .. controls (314.2,221.3) and (324.05,211.34) .. (336.2,211.34) .. controls (348.35,211.34) and (358.2,221.3) .. (358.2,233.59) .. controls (358.2,245.88) and (348.35,255.84) .. (336.2,255.84) .. controls (324.05,255.84) and (314.2,245.88) .. (314.2,233.59) -- cycle ;
\draw [line width=1.5]    (257,169.5) -- (332.66,209.47) ;
\draw [shift={(336.2,211.34)}, rotate = 207.84] [fill={rgb, 255:red, 0; green, 0; blue, 0 }  ][line width=0.08]  [draw opacity=0] (13.4,-6.43) -- (0,0) -- (13.4,6.44) -- (8.9,0) -- cycle    ;
\draw [line width=1.5]    (336.2,169.5) -- (336.2,207.34) ;
\draw [shift={(336.2,211.34)}, rotate = 270] [fill={rgb, 255:red, 0; green, 0; blue, 0 }  ][line width=0.08]  [draw opacity=0] (13.4,-6.43) -- (0,0) -- (13.4,6.44) -- (8.9,0) -- cycle    ;
\draw [line width=1.5]    (415.4,169.5) -- (339.74,209.47) ;
\draw [shift={(336.2,211.34)}, rotate = 332.16] [fill={rgb, 255:red, 0; green, 0; blue, 0 }  ][line width=0.08]  [draw opacity=0] (13.4,-6.43) -- (0,0) -- (13.4,6.44) -- (8.9,0) -- cycle    ;
\draw [line width=1.5]    (27,169.5) .. controls (99.93,199.88) and (137.72,186.25) .. (181.98,170.7) ;
\draw [shift={(185.4,169.5)}, rotate = 160.7] [fill={rgb, 255:red, 0; green, 0; blue, 0 }  ][line width=0.08]  [draw opacity=0] (13.4,-6.43) -- (0,0) -- (13.4,6.44) -- (8.9,0) -- cycle    ;
\draw [line width=1.5]    (49,147.25) -- (80.2,147.25) ;
\draw [shift={(84.2,147.25)}, rotate = 180] [fill={rgb, 255:red, 0; green, 0; blue, 0 }  ][line width=0.08]  [draw opacity=0] (13.4,-6.43) -- (0,0) -- (13.4,6.44) -- (8.9,0) -- cycle    ;
\draw  [line width=1.5]  (460,147.25) .. controls (460,134.96) and (469.85,125) .. (482,125) .. controls (494.15,125) and (504,134.96) .. (504,147.25) .. controls (504,159.54) and (494.15,169.5) .. (482,169.5) .. controls (469.85,169.5) and (460,159.54) .. (460,147.25) -- cycle ;
\draw  [line width=1.5]  (618.4,147.25) .. controls (618.4,134.96) and (628.25,125) .. (640.4,125) .. controls (652.55,125) and (662.4,134.96) .. (662.4,147.25) .. controls (662.4,159.54) and (652.55,169.5) .. (640.4,169.5) .. controls (628.25,169.5) and (618.4,159.54) .. (618.4,147.25) -- cycle ;
\draw  [line width=1.5]  (539.2,147.25) .. controls (539.2,134.96) and (549.05,125) .. (561.2,125) .. controls (573.35,125) and (583.2,134.96) .. (583.2,147.25) .. controls (583.2,159.54) and (573.35,169.5) .. (561.2,169.5) .. controls (549.05,169.5) and (539.2,159.54) .. (539.2,147.25) -- cycle ;
\draw  [line width=1.5]  (539.2,233.59) .. controls (539.2,221.3) and (549.05,211.34) .. (561.2,211.34) .. controls (573.35,211.34) and (583.2,221.3) .. (583.2,233.59) .. controls (583.2,245.88) and (573.35,255.84) .. (561.2,255.84) .. controls (549.05,255.84) and (539.2,245.88) .. (539.2,233.59) -- cycle ;
\draw [line width=1.5]    (482,169.5) -- (557.66,209.47) ;
\draw [shift={(561.2,211.34)}, rotate = 207.84] [fill={rgb, 255:red, 0; green, 0; blue, 0 }  ][line width=0.08]  [draw opacity=0] (13.4,-6.43) -- (0,0) -- (13.4,6.44) -- (8.9,0) -- cycle    ;
\draw [line width=1.5]    (561.2,169.5) -- (561.2,207.34) ;
\draw [shift={(561.2,211.34)}, rotate = 270] [fill={rgb, 255:red, 0; green, 0; blue, 0 }  ][line width=0.08]  [draw opacity=0] (13.4,-6.43) -- (0,0) -- (13.4,6.44) -- (8.9,0) -- cycle    ;
\draw [line width=1.5]    (640.4,169.5) -- (564.74,209.47) ;
\draw [shift={(561.2,211.34)}, rotate = 332.16] [fill={rgb, 255:red, 0; green, 0; blue, 0 }  ][line width=0.08]  [draw opacity=0] (13.4,-6.43) -- (0,0) -- (13.4,6.44) -- (8.9,0) -- cycle    ;
\draw [color={rgb, 255:red, 0; green, 0; blue, 255 }  ,draw opacity=1 ][line width=1.5]    (482,169.5) .. controls (554.93,199.88) and (592.72,186.25) .. (636.98,170.7) ;
\draw [shift={(640.4,169.5)}, rotate = 160.7] [fill={rgb, 255:red, 0; green, 0; blue, 255 }  ,fill opacity=1 ][line width=0.08]  [draw opacity=0] (13.4,-6.43) -- (0,0) -- (13.4,6.44) -- (8.9,0) -- cycle    ;
\draw [color={rgb, 255:red, 0; green, 0; blue, 255 }  ,draw opacity=1 ][line width=1.5]    (504,147.25) -- (535.2,147.25) ;
\draw [shift={(539.2,147.25)}, rotate = 180] [fill={rgb, 255:red, 0; green, 0; blue, 255 }  ,fill opacity=1 ][line width=0.08]  [draw opacity=0] (13.4,-6.43) -- (0,0) -- (13.4,6.44) -- (8.9,0) -- cycle    ;

\draw (13.84,139.05) node [anchor=north west][inner sep=0.75pt]  [font=\LARGE]  {$X_{1}$};
\draw (93.34,138.16) node [anchor=north west][inner sep=0.75pt]  [font=\LARGE]  {$X_{2}$};
\draw (170.86,139.05) node [anchor=north west][inner sep=0.75pt]  [font=\LARGE]  {$X_{3}$};
\draw (243.84,139.05) node [anchor=north west][inner sep=0.75pt]  [font=\LARGE]  {$X_{1}$};
\draw (323.34,138.16) node [anchor=north west][inner sep=0.75pt]  [font=\LARGE]  {$X_{2}$};
\draw (401.45,139.05) node [anchor=north west][inner sep=0.75pt]  [font=\LARGE]  {$X_{3}$};
\draw (328.93,223.83) node [anchor=north west][inner sep=0.75pt]  [font=\LARGE]  {$\hat{Y}$};
\draw (468.84,139.05) node [anchor=north west][inner sep=0.75pt]  [font=\LARGE]  {$X_{1}$};
\draw (548.34,138.16) node [anchor=north west][inner sep=0.75pt]  [font=\LARGE]  {$X_{2}$};
\draw (625.86,139.05) node [anchor=north west][inner sep=0.75pt]  [font=\LARGE]  {$X_{3}$};
\draw (98.93,224.83) node [anchor=north west][inner sep=0.75pt]  [font=\LARGE]  {$Y$};
\draw (554.93,223.83) node [anchor=north west][inner sep=0.75pt]  [font=\LARGE]  {$\hat{Y}$};
\draw (96,266.4) node [anchor=north west][inner sep=0.75pt]  [font=\LARGE]  {$\mathcal{G}$};
\draw (328,266.4) node [anchor=north west][inner sep=0.75pt]  [font=\LARGE]  {$\mathcal{N}$};
\draw (550,266.4) node [anchor=north west][inner sep=0.75pt]  [font=\LARGE]  {$\mathcal{N}^{Ind}$};

\end{tikzpicture}

%% file: images/synthetic.tikz
\begin{tikzpicture}[x=0.75pt,y=0.75pt,yscale=-1,xscale=1]

\draw  [line width=1.5]  (350,41) .. controls (350,27.19) and (361.19,16) .. (375,16) .. controls (388.81,16) and (400,27.19) .. (400,41) .. controls (400,54.81) and (388.81,66) .. (375,66) .. controls (361.19,66) and (350,54.81) .. (350,41) -- cycle ;
\draw  [line width=1.5]  (250,91) .. controls (250,77.19) and (261.19,66) .. (275,66) .. controls (288.81,66) and (300,77.19) .. (300,91) .. controls (300,104.81) and (288.81,116) .. (275,116) .. controls (261.19,116) and (250,104.81) .. (250,91) -- cycle ;
\draw [line width=1.5]    (300,91) -- (347.17,43.83) ;
\draw [shift={(350,41)}, rotate = 135] [fill={rgb, 255:red, 0; green, 0; blue, 0 }  ][line width=0.08]  [draw opacity=0] (13.4,-6.43) -- (0,0) -- (13.4,6.44) -- (8.9,0) -- cycle    ;
\draw  [line width=1.5]  (350,141) .. controls (350,127.19) and (361.19,116) .. (375,116) .. controls (388.81,116) and (400,127.19) .. (400,141) .. controls (400,154.81) and (388.81,166) .. (375,166) .. controls (361.19,166) and (350,154.81) .. (350,141) -- cycle ;
\draw [line width=1.5]    (300,91) -- (347.17,138.17) ;
\draw [shift={(350,141)}, rotate = 225] [fill={rgb, 255:red, 0; green, 0; blue, 0 }  ][line width=0.08]  [draw opacity=0] (13.4,-6.43) -- (0,0) -- (13.4,6.44) -- (8.9,0) -- cycle    ;
\draw  [line width=1.5]  (450,91) .. controls (450,77.19) and (461.19,66) .. (475,66) .. controls (488.81,66) and (500,77.19) .. (500,91) .. controls (500,104.81) and (488.81,116) .. (475,116) .. controls (461.19,116) and (450,104.81) .. (450,91) -- cycle ;
\draw [line width=1.5]    (375,66) -- (375,112) ;
\draw [shift={(375,116)}, rotate = 270] [fill={rgb, 255:red, 0; green, 0; blue, 0 }  ][line width=0.08]  [draw opacity=0] (13.4,-6.43) -- (0,0) -- (13.4,6.44) -- (8.9,0) -- cycle    ;
\draw [line width=1.5]    (400,141) -- (447.17,93.83) ;
\draw [shift={(450,91)}, rotate = 135] [fill={rgb, 255:red, 0; green, 0; blue, 0 }  ][line width=0.08]  [draw opacity=0] (13.4,-6.43) -- (0,0) -- (13.4,6.44) -- (8.9,0) -- cycle    ;
\draw [line width=1.5]    (400,41) -- (447.17,88.17) ;
\draw [shift={(450,91)}, rotate = 225] [fill={rgb, 255:red, 0; green, 0; blue, 0 }  ][line width=0.08]  [draw opacity=0] (13.4,-6.43) -- (0,0) -- (13.4,6.44) -- (8.9,0) -- cycle    ;

\draw (365,32) node [anchor=north west][inner sep=0.75pt]  [font=\huge] [align=left] {Z};
\draw (263,82) node [anchor=north west][inner sep=0.75pt]  [font=\huge] [align=left] {W};
\draw (365,132) node [anchor=north west][inner sep=0.75pt]  [font=\huge] [align=left] {X};
\draw (465,82) node [anchor=north west][inner sep=0.75pt]  [font=\huge] [align=left] {Y};

\end{tikzpicture}

%% file: images/autompg_asia.tikz
\begin{tikzpicture}[x=0.75pt,y=0.75pt,yscale=-1,xscale=1]

\draw  [line width=1.5]  (5,79.5) .. controls (5,65.69) and (16.19,54.5) .. (30,54.5) .. controls (43.81,54.5) and (55,65.69) .. (55,79.5) .. controls (55,93.31) and (43.81,104.5) .. (30,104.5) .. controls (16.19,104.5) and (5,93.31) .. (5,79.5) -- cycle ;
\draw  [line width=1.5]  (105,29.5) .. controls (105,15.69) and (116.19,4.5) .. (130,4.5) .. controls (143.81,4.5) and (155,15.69) .. (155,29.5) .. controls (155,43.31) and (143.81,54.5) .. (130,54.5) .. controls (116.19,54.5) and (105,43.31) .. (105,29.5) -- cycle ;
\draw  [line width=1.5]  (105,129.5) .. controls (105,115.69) and (116.19,104.5) .. (130,104.5) .. controls (143.81,104.5) and (155,115.69) .. (155,129.5) .. controls (155,143.31) and (143.81,154.5) .. (130,154.5) .. controls (116.19,154.5) and (105,143.31) .. (105,129.5) -- cycle ;
\draw  [line width=1.5]  (205,129.5) .. controls (205,115.69) and (216.19,104.5) .. (230,104.5) .. controls (243.81,104.5) and (255,115.69) .. (255,129.5) .. controls (255,143.31) and (243.81,154.5) .. (230,154.5) .. controls (216.19,154.5) and (205,143.31) .. (205,129.5) -- cycle ;
\draw  [line width=1.5]  (205,54.5) .. controls (205,40.69) and (216.19,29.5) .. (230,29.5) .. controls (243.81,29.5) and (255,40.69) .. (255,54.5) .. controls (255,68.31) and (243.81,79.5) .. (230,79.5) .. controls (216.19,79.5) and (205,68.31) .. (205,54.5) -- cycle ;
\draw  [line width=1.5]  (305,79.5) .. controls (305,65.69) and (316.19,54.5) .. (330,54.5) .. controls (343.81,54.5) and (355,65.69) .. (355,79.5) .. controls (355,93.31) and (343.81,104.5) .. (330,104.5) .. controls (316.19,104.5) and (305,93.31) .. (305,79.5) -- cycle ;
\draw [line width=1.5]    (55,79.5) -- (102.17,32.33) ;
\draw [shift={(105,29.5)}, rotate = 135] [fill={rgb, 255:red, 0; green, 0; blue, 0 }  ][line width=0.08]  [draw opacity=0] (13.4,-6.43) -- (0,0) -- (13.4,6.44) -- (8.9,0) -- cycle    ;
\draw [line width=1.5]    (55,79.5) -- (102.17,126.67) ;
\draw [shift={(105,129.5)}, rotate = 225] [fill={rgb, 255:red, 0; green, 0; blue, 0 }  ][line width=0.08]  [draw opacity=0] (13.4,-6.43) -- (0,0) -- (13.4,6.44) -- (8.9,0) -- cycle    ;
\draw [line width=1.5]    (154.5,36.5) -- (200.63,48.49) ;
\draw [shift={(204.5,49.5)}, rotate = 194.57] [fill={rgb, 255:red, 0; green, 0; blue, 0 }  ][line width=0.08]  [draw opacity=0] (13.4,-6.43) -- (0,0) -- (13.4,6.44) -- (8.9,0) -- cycle    ;
\draw [line width=1.5]    (230,104.5) -- (230,83.5) ;
\draw [shift={(230,79.5)}, rotate = 90] [fill={rgb, 255:red, 0; green, 0; blue, 0 }  ][line width=0.08]  [draw opacity=0] (13.4,-6.43) -- (0,0) -- (13.4,6.44) -- (8.9,0) -- cycle    ;
\draw [line width=1.5]    (254.5,49.5) -- (301.56,77.46) ;
\draw [shift={(305,79.5)}, rotate = 210.71] [fill={rgb, 255:red, 0; green, 0; blue, 0 }  ][line width=0.08]  [draw opacity=0] (13.4,-6.43) -- (0,0) -- (13.4,6.44) -- (8.9,0) -- cycle    ;
\draw [line width=1.5]    (255,129.5) -- (302.17,82.33) ;
\draw [shift={(305,79.5)}, rotate = 135] [fill={rgb, 255:red, 0; green, 0; blue, 0 }  ][line width=0.08]  [draw opacity=0] (13.4,-6.43) -- (0,0) -- (13.4,6.44) -- (8.9,0) -- cycle    ;
\draw [line width=1.5]    (155,129.5) -- (201,129.5) ;
\draw [shift={(205,129.5)}, rotate = 180] [fill={rgb, 255:red, 0; green, 0; blue, 0 }  ][line width=0.08]  [draw opacity=0] (13.4,-6.43) -- (0,0) -- (13.4,6.44) -- (8.9,0) -- cycle    ;
\draw [line width=1.5]    (55,79.5) .. controls (106.98,221.32) and (275.58,194.31) .. (304.18,82.9) ;
\draw [shift={(305,79.5)}, rotate = 102.82] [fill={rgb, 255:red, 0; green, 0; blue, 0 }  ][line width=0.08]  [draw opacity=0] (13.4,-6.43) -- (0,0) -- (13.4,6.44) -- (8.9,0) -- cycle    ;
\draw [line width=1.5]    (154.5,36.5) .. controls (187.01,10.64) and (270.45,-0.42) .. (303.52,75.95) ;
\draw [shift={(305,79.5)}, rotate = 248.26] [fill={rgb, 255:red, 0; green, 0; blue, 0 }  ][line width=0.08]  [draw opacity=0] (13.4,-6.43) -- (0,0) -- (13.4,6.44) -- (8.9,0) -- cycle    ;
\draw [line width=1.5]    (155,129.5) -- (202.4,52.9) ;
\draw [shift={(204.5,49.5)}, rotate = 121.75] [fill={rgb, 255:red, 0; green, 0; blue, 0 }  ][line width=0.08]  [draw opacity=0] (13.4,-6.43) -- (0,0) -- (13.4,6.44) -- (8.9,0) -- cycle    ;
\draw [line width=1.5]    (55,79.5) .. controls (99.16,72.54) and (182.95,101.75) .. (202.85,126.44) ;
\draw [shift={(205,129.5)}, rotate = 239.29] [fill={rgb, 255:red, 0; green, 0; blue, 0 }  ][line width=0.08]  [draw opacity=0] (13.4,-6.43) -- (0,0) -- (13.4,6.44) -- (8.9,0) -- cycle    ;
\draw  [line width=1.5]  (460,45) .. controls (460,31.19) and (471.19,20) .. (485,20) .. controls (498.81,20) and (510,31.19) .. (510,45) .. controls (510,58.81) and (498.81,70) .. (485,70) .. controls (471.19,70) and (460,58.81) .. (460,45) -- cycle ;
\draw  [line width=1.5]  (460,145) .. controls (460,131.19) and (471.19,120) .. (485,120) .. controls (498.81,120) and (510,131.19) .. (510,145) .. controls (510,158.81) and (498.81,170) .. (485,170) .. controls (471.19,170) and (460,158.81) .. (460,145) -- cycle ;
\draw  [line width=1.5]  (540,45) .. controls (540,31.19) and (551.19,20) .. (565,20) .. controls (578.81,20) and (590,31.19) .. (590,45) .. controls (590,58.81) and (578.81,70) .. (565,70) .. controls (551.19,70) and (540,58.81) .. (540,45) -- cycle ;
\draw  [line width=1.5]  (540,145) .. controls (540,131.19) and (551.19,120) .. (565,120) .. controls (578.81,120) and (590,131.19) .. (590,145) .. controls (590,158.81) and (578.81,170) .. (565,170) .. controls (551.19,170) and (540,158.81) .. (540,145) -- cycle ;
\draw  [line width=1.5]  (620,45) .. controls (620,31.19) and (631.19,20) .. (645,20) .. controls (658.81,20) and (670,31.19) .. (670,45) .. controls (670,58.81) and (658.81,70) .. (645,70) .. controls (631.19,70) and (620,58.81) .. (620,45) -- cycle ;
\draw  [line width=1.5]  (620,145) .. controls (620,131.19) and (631.19,120) .. (645,120) .. controls (658.81,120) and (670,131.19) .. (670,145) .. controls (670,158.81) and (658.81,170) .. (645,170) .. controls (631.19,170) and (620,158.81) .. (620,145) -- cycle ;
\draw  [line width=1.5]  (380,45) .. controls (380,31.19) and (391.19,20) .. (405,20) .. controls (418.81,20) and (430,31.19) .. (430,45) .. controls (430,58.81) and (418.81,70) .. (405,70) .. controls (391.19,70) and (380,58.81) .. (380,45) -- cycle ;
\draw  [line width=1.5]  (380,145) .. controls (380,131.19) and (391.19,120) .. (405,120) .. controls (418.81,120) and (430,131.19) .. (430,145) .. controls (430,158.81) and (418.81,170) .. (405,170) .. controls (391.19,170) and (380,158.81) .. (380,145) -- cycle ;
\draw [line width=1.5]    (430,45) -- (456,45) ;
\draw [shift={(460,45)}, rotate = 180] [fill={rgb, 255:red, 0; green, 0; blue, 0 }  ][line width=0.08]  [draw opacity=0] (13.4,-6.43) -- (0,0) -- (13.4,6.44) -- (8.9,0) -- cycle    ;
\draw [line width=1.5]    (510,45) -- (536,45) ;
\draw [shift={(540,45)}, rotate = 180] [fill={rgb, 255:red, 0; green, 0; blue, 0 }  ][line width=0.08]  [draw opacity=0] (13.4,-6.43) -- (0,0) -- (13.4,6.44) -- (8.9,0) -- cycle    ;
\draw [line width=1.5]    (430,145) -- (456,145) ;
\draw [shift={(460,145)}, rotate = 180] [fill={rgb, 255:red, 0; green, 0; blue, 0 }  ][line width=0.08]  [draw opacity=0] (13.4,-6.43) -- (0,0) -- (13.4,6.44) -- (8.9,0) -- cycle    ;
\draw [line width=1.5]    (510,145) -- (538.85,48.83) ;
\draw [shift={(540,45)}, rotate = 106.7] [fill={rgb, 255:red, 0; green, 0; blue, 0 }  ][line width=0.08]  [draw opacity=0] (13.4,-6.43) -- (0,0) -- (13.4,6.44) -- (8.9,0) -- cycle    ;
\draw [line width=1.5]    (590,45) -- (616,45) ;
\draw [shift={(620,45)}, rotate = 180] [fill={rgb, 255:red, 0; green, 0; blue, 0 }  ][line width=0.08]  [draw opacity=0] (13.4,-6.43) -- (0,0) -- (13.4,6.44) -- (8.9,0) -- cycle    ;
\draw [line width=1.5]    (590,45) -- (618.85,141.17) ;
\draw [shift={(620,145)}, rotate = 253.3] [fill={rgb, 255:red, 0; green, 0; blue, 0 }  ][line width=0.08]  [draw opacity=0] (13.4,-6.43) -- (0,0) -- (13.4,6.44) -- (8.9,0) -- cycle    ;
\draw [line width=1.5]    (590,145) -- (616,145) ;
\draw [shift={(620,145)}, rotate = 180] [fill={rgb, 255:red, 0; green, 0; blue, 0 }  ][line width=0.08]  [draw opacity=0] (13.4,-6.43) -- (0,0) -- (13.4,6.44) -- (8.9,0) -- cycle    ;
\draw [line width=1.5]    (430,145) .. controls (469.4,115.45) and (478.24,89.54) .. (537.27,142.52) ;
\draw [shift={(540,145)}, rotate = 222.45] [fill={rgb, 255:red, 0; green, 0; blue, 0 }  ][line width=0.08]  [draw opacity=0] (13.4,-6.43) -- (0,0) -- (13.4,6.44) -- (8.9,0) -- cycle    ;
\draw  [line width=1.5]  (110.54,471.54) .. controls (96.73,471.54) and (85.54,460.34) .. (85.54,446.54) .. controls (85.54,432.73) and (96.73,421.54) .. (110.54,421.54) .. controls (124.34,421.54) and (135.54,432.73) .. (135.54,446.54) .. controls (135.54,460.34) and (124.34,471.54) .. (110.54,471.54) -- cycle ;
\draw [line width=1.5]    (135.54,446.54) -- (222.55,369.19) ;
\draw [shift={(225.54,366.54)}, rotate = 138.37] [fill={rgb, 255:red, 0; green, 0; blue, 0 }  ][line width=0.08]  [draw opacity=0] (13.4,-6.43) -- (0,0) -- (13.4,6.44) -- (8.9,0) -- cycle    ;
\draw [line width=1.5]    (135.54,446.54) -- (181.54,446.54) ;
\draw [shift={(185.54,446.54)}, rotate = 180] [fill={rgb, 255:red, 0; green, 0; blue, 0 }  ][line width=0.08]  [draw opacity=0] (13.4,-6.43) -- (0,0) -- (13.4,6.44) -- (8.9,0) -- cycle    ;
\draw  [line width=1.5]  (210.54,471.54) .. controls (196.73,471.54) and (185.54,460.34) .. (185.54,446.54) .. controls (185.54,432.73) and (196.73,421.54) .. (210.54,421.54) .. controls (224.34,421.54) and (235.54,432.73) .. (235.54,446.54) .. controls (235.54,460.34) and (224.34,471.54) .. (210.54,471.54) -- cycle ;
\draw  [line width=1.5]  (250.54,391.54) .. controls (236.73,391.54) and (225.54,380.34) .. (225.54,366.54) .. controls (225.54,352.73) and (236.73,341.54) .. (250.54,341.54) .. controls (264.34,341.54) and (275.54,352.73) .. (275.54,366.54) .. controls (275.54,380.34) and (264.34,391.54) .. (250.54,391.54) -- cycle ;
\draw  [line width=1.5]  (250.54,551.54) .. controls (236.73,551.54) and (225.54,540.34) .. (225.54,526.54) .. controls (225.54,512.73) and (236.73,501.54) .. (250.54,501.54) .. controls (264.34,501.54) and (275.54,512.73) .. (275.54,526.54) .. controls (275.54,540.34) and (264.34,551.54) .. (250.54,551.54) -- cycle ;
\draw [line width=1.5]    (135.54,446.54) -- (222.55,523.88) ;
\draw [shift={(225.54,526.54)}, rotate = 221.63] [fill={rgb, 255:red, 0; green, 0; blue, 0 }  ][line width=0.08]  [draw opacity=0] (13.4,-6.43) -- (0,0) -- (13.4,6.44) -- (8.9,0) -- cycle    ;
\draw  [line width=1.5]  (250.54,311.54) .. controls (236.73,311.54) and (225.54,300.34) .. (225.54,286.54) .. controls (225.54,272.73) and (236.73,261.54) .. (250.54,261.54) .. controls (264.34,261.54) and (275.54,272.73) .. (275.54,286.54) .. controls (275.54,300.34) and (264.34,311.54) .. (250.54,311.54) -- cycle ;
\draw [line width=1.5]    (135.54,446.54) -- (223.58,290.02) ;
\draw [shift={(225.54,286.54)}, rotate = 119.36] [fill={rgb, 255:red, 0; green, 0; blue, 0 }  ][line width=0.08]  [draw opacity=0] (13.4,-6.43) -- (0,0) -- (13.4,6.44) -- (8.9,0) -- cycle    ;
\draw  [line width=1.5]  (350.54,511.54) .. controls (336.73,511.54) and (325.54,500.34) .. (325.54,486.54) .. controls (325.54,472.73) and (336.73,461.54) .. (350.54,461.54) .. controls (364.34,461.54) and (375.54,472.73) .. (375.54,486.54) .. controls (375.54,500.34) and (364.34,511.54) .. (350.54,511.54) -- cycle ;
\draw  [line width=1.5]  (450.54,471.54) .. controls (436.73,471.54) and (425.54,460.34) .. (425.54,446.54) .. controls (425.54,432.73) and (436.73,421.54) .. (450.54,421.54) .. controls (464.34,421.54) and (475.54,432.73) .. (475.54,446.54) .. controls (475.54,460.34) and (464.34,471.54) .. (450.54,471.54) -- cycle ;
\draw  [line width=1.5]  (550.54,431.54) .. controls (536.73,431.54) and (525.54,420.34) .. (525.54,406.54) .. controls (525.54,392.73) and (536.73,381.54) .. (550.54,381.54) .. controls (564.34,381.54) and (575.54,392.73) .. (575.54,406.54) .. controls (575.54,420.34) and (564.34,431.54) .. (550.54,431.54) -- cycle ;
\draw [line width=1.5]    (275.54,526.54) -- (322.41,489.04) ;
\draw [shift={(325.54,486.54)}, rotate = 141.34] [fill={rgb, 255:red, 0; green, 0; blue, 0 }  ][line width=0.08]  [draw opacity=0] (13.4,-6.43) -- (0,0) -- (13.4,6.44) -- (8.9,0) -- cycle    ;
\draw [line width=1.5]    (235.54,446.54) -- (321.88,484.91) ;
\draw [shift={(325.54,486.54)}, rotate = 203.96] [fill={rgb, 255:red, 0; green, 0; blue, 0 }  ][line width=0.08]  [draw opacity=0] (13.4,-6.43) -- (0,0) -- (13.4,6.44) -- (8.9,0) -- cycle    ;
\draw [line width=1.5]    (375.54,486.54) -- (422.41,449.04) ;
\draw [shift={(425.54,446.54)}, rotate = 141.34] [fill={rgb, 255:red, 0; green, 0; blue, 0 }  ][line width=0.08]  [draw opacity=0] (13.4,-6.43) -- (0,0) -- (13.4,6.44) -- (8.9,0) -- cycle    ;
\draw [line width=1.5]    (475.54,446.54) -- (522.41,409.04) ;
\draw [shift={(525.54,406.54)}, rotate = 141.34] [fill={rgb, 255:red, 0; green, 0; blue, 0 }  ][line width=0.08]  [draw opacity=0] (13.4,-6.43) -- (0,0) -- (13.4,6.44) -- (8.9,0) -- cycle    ;
\draw [line width=1.5]    (235.54,446.54) -- (421.54,446.54) ;
\draw [shift={(425.54,446.54)}, rotate = 180] [fill={rgb, 255:red, 0; green, 0; blue, 0 }  ][line width=0.08]  [draw opacity=0] (13.4,-6.43) -- (0,0) -- (13.4,6.44) -- (8.9,0) -- cycle    ;
\draw [line width=1.5]    (235.54,446.54) -- (521.57,407.08) ;
\draw [shift={(525.54,406.54)}, rotate = 172.15] [fill={rgb, 255:red, 0; green, 0; blue, 0 }  ][line width=0.08]  [draw opacity=0] (13.4,-6.43) -- (0,0) -- (13.4,6.44) -- (8.9,0) -- cycle    ;
\draw [line width=1.5]    (235.54,446.54) -- (249.48,497.68) ;
\draw [shift={(250.54,501.54)}, rotate = 254.74] [fill={rgb, 255:red, 0; green, 0; blue, 0 }  ][line width=0.08]  [draw opacity=0] (13.4,-6.43) -- (0,0) -- (13.4,6.44) -- (8.9,0) -- cycle    ;
\draw [line width=1.5]    (235.54,446.54) -- (249.48,395.4) ;
\draw [shift={(250.54,391.54)}, rotate = 105.26] [fill={rgb, 255:red, 0; green, 0; blue, 0 }  ][line width=0.08]  [draw opacity=0] (13.4,-6.43) -- (0,0) -- (13.4,6.44) -- (8.9,0) -- cycle    ;
\draw [line width=1.5]    (235.54,446.54) .. controls (267.55,403.2) and (324.3,355.98) .. (253.85,313.48) ;
\draw [shift={(250.54,311.54)}, rotate = 29.66] [fill={rgb, 255:red, 0; green, 0; blue, 0 }  ][line width=0.08]  [draw opacity=0] (13.4,-6.43) -- (0,0) -- (13.4,6.44) -- (8.9,0) -- cycle    ;
\draw [line width=1.5]    (135.54,446.54) .. controls (235.47,649.34) and (315.92,536.53) .. (324.98,489.98) ;
\draw [shift={(325.54,486.54)}, rotate = 97.13] [fill={rgb, 255:red, 0; green, 0; blue, 0 }  ][line width=0.08]  [draw opacity=0] (13.4,-6.43) -- (0,0) -- (13.4,6.44) -- (8.9,0) -- cycle    ;
\draw  [line width=1.5]  (338.54,371.54) .. controls (324.73,371.54) and (313.54,360.34) .. (313.54,346.54) .. controls (313.54,332.73) and (324.73,321.54) .. (338.54,321.54) .. controls (352.34,321.54) and (363.54,332.73) .. (363.54,346.54) .. controls (363.54,360.34) and (352.34,371.54) .. (338.54,371.54) -- cycle ;
\draw  [line width=1.5]  (438.54,331.54) .. controls (424.73,331.54) and (413.54,320.34) .. (413.54,306.54) .. controls (413.54,292.73) and (424.73,281.54) .. (438.54,281.54) .. controls (452.34,281.54) and (463.54,292.73) .. (463.54,306.54) .. controls (463.54,320.34) and (452.34,331.54) .. (438.54,331.54) -- cycle ;
\draw  [line width=1.5]  (538.54,371.54) .. controls (524.73,371.54) and (513.54,360.34) .. (513.54,346.54) .. controls (513.54,332.73) and (524.73,321.54) .. (538.54,321.54) .. controls (552.34,321.54) and (563.54,332.73) .. (563.54,346.54) .. controls (563.54,360.34) and (552.34,371.54) .. (538.54,371.54) -- cycle ;
\draw [line width=1.5]    (363.54,346.54) -- (410.41,309.04) ;
\draw [shift={(413.54,306.54)}, rotate = 141.34] [fill={rgb, 255:red, 0; green, 0; blue, 0 }  ][line width=0.08]  [draw opacity=0] (13.4,-6.43) -- (0,0) -- (13.4,6.44) -- (8.9,0) -- cycle    ;
\draw [line width=1.5]    (463.54,306.54) -- (510.41,344.04) ;
\draw [shift={(513.54,346.54)}, rotate = 218.66] [fill={rgb, 255:red, 0; green, 0; blue, 0 }  ][line width=0.08]  [draw opacity=0] (13.4,-6.43) -- (0,0) -- (13.4,6.44) -- (8.9,0) -- cycle    ;
\draw [line width=1.5]    (363.54,346.54) -- (509.54,346.54) ;
\draw [shift={(513.54,346.54)}, rotate = 180] [fill={rgb, 255:red, 0; green, 0; blue, 0 }  ][line width=0.08]  [draw opacity=0] (13.4,-6.43) -- (0,0) -- (13.4,6.44) -- (8.9,0) -- cycle    ;

\draw (18,70.5) node [anchor=north west][inner sep=0.75pt]  [font=\huge] [align=left] {C};
\draw (118,18.5) node [anchor=north west][inner sep=0.75pt]  [font=\huge] [align=left] {W};
\draw (120,120.5) node [anchor=north west][inner sep=0.75pt]  [font=\huge] [align=left] {D};
\draw (220,120.5) node [anchor=north west][inner sep=0.75pt]  [font=\huge] [align=left] {H};
\draw (219,43.5) node [anchor=north west][inner sep=0.75pt]  [font=\huge] [align=left] {A};
\draw (319,69.5) node [anchor=north west][inner sep=0.75pt]  [font=\huge] [align=left] {M};
\draw (475,36) node [anchor=north west][inner sep=0.75pt]  [font=\huge] [align=left] {T};
\draw (475,136) node [anchor=north west][inner sep=0.75pt]  [font=\huge] [align=left] {L};
\draw (553,34) node [anchor=north west][inner sep=0.75pt]  [font=\huge] [align=left] {E};
\draw (555,136) node [anchor=north west][inner sep=0.75pt]  [font=\huge] [align=left] {B};
\draw (635,36) node [anchor=north west][inner sep=0.75pt]  [font=\huge] [align=left] {X};
\draw (635,136) node [anchor=north west][inner sep=0.75pt]  [font=\huge] [align=left] {D};
\draw (395,34) node [anchor=north west][inner sep=0.75pt]  [font=\huge] [align=left] {A};
\draw (395,136) node [anchor=north west][inner sep=0.75pt]  [font=\huge] [align=left] {S};
\draw (108,193) node [anchor=north west][inner sep=0.75pt]  [font=\Huge] [align=left] {(i) Auto-MPG};
\draw (449,193) node [anchor=north west][inner sep=0.75pt]  [font=\Huge] [align=left] {(ii) Lung Cancer};
\draw (90,437) node [anchor=north west][inner sep=0.75pt]  [font=\LARGE] [align=left] {PKC};
\draw (232,279) node [anchor=north west][inner sep=0.75pt]  [font=\LARGE] [align=left] {P38};
\draw (234,359) node [anchor=north west][inner sep=0.75pt]  [font=\LARGE] [align=left] {Jnk};
\draw (191,439) node [anchor=north west][inner sep=0.75pt]  [font=\LARGE] [align=left] {PKA};
\draw (234,518) node [anchor=north west][inner sep=0.75pt]  [font=\LARGE] [align=left] {Raf};
\draw (331,479) node [anchor=north west][inner sep=0.75pt]  [font=\LARGE] [align=left] {Mek};
\draw (320,338) node [anchor=north west][inner sep=0.75pt]  [font=\LARGE] [align=left] {Plcg};
\draw (419,299) node [anchor=north west][inner sep=0.75pt]  [font=\LARGE] [align=left] {PIP3};
\draw (520,339) node [anchor=north west][inner sep=0.75pt]  [font=\LARGE] [align=left] {PIP2};
\draw (435,439) node [anchor=north west][inner sep=0.75pt]  [font=\LARGE] [align=left] {Erk};
\draw (534,398) node [anchor=north west][inner sep=0.75pt]  [font=\LARGE] [align=left] {Akt};
\draw (303,568) node [anchor=north west][inner sep=0.75pt]  [font=\Huge] [align=left] {(iii) Sachs};

\end{tikzpicture}

%% file: images/direct_indirect.tikz
\tikzset{every picture/.style={line width=0.75pt}} 

\begin{tikzpicture}[x=0.75pt,y=0.75pt,yscale=-1,xscale=1]

\draw  [line width=1.5]  (75,220) .. controls (61.19,220) and (50,208.81) .. (50,195) .. controls (50,181.19) and (61.19,170) .. (75,170) .. controls (88.81,170) and (100,181.19) .. (100,195) .. controls (100,208.81) and (88.81,220) .. (75,220) -- cycle ;
\draw [color={rgb, 255:red, 255; green, 0; blue, 0 }  ,draw opacity=1 ][line width=1.5]    (100,195) -- (146,195) ;
\draw [shift={(150,195)}, rotate = 180] [fill={rgb, 255:red, 255; green, 0; blue, 0 }  ,fill opacity=1 ][line width=0.08]  [draw opacity=0] (13.4,-6.43) -- (0,0) -- (13.4,6.44) -- (8.9,0) -- cycle    ;
\draw  [line width=1.5]  (175,220) .. controls (161.19,220) and (150,208.81) .. (150,195) .. controls (150,181.19) and (161.19,170) .. (175,170) .. controls (188.81,170) and (200,181.19) .. (200,195) .. controls (200,208.81) and (188.81,220) .. (175,220) -- cycle ;
\draw [color={rgb, 255:red, 255; green, 0; blue, 0 }  ,draw opacity=1 ][line width=1.5]    (200,195) -- (246,195) ;
\draw [shift={(250,195)}, rotate = 180] [fill={rgb, 255:red, 255; green, 0; blue, 0 }  ,fill opacity=1 ][line width=0.08]  [draw opacity=0] (13.4,-6.43) -- (0,0) -- (13.4,6.44) -- (8.9,0) -- cycle    ;
\draw  [line width=1.5]  (275,220) .. controls (261.19,220) and (250,208.81) .. (250,195) .. controls (250,181.19) and (261.19,170) .. (275,170) .. controls (288.81,170) and (300,181.19) .. (300,195) .. controls (300,208.81) and (288.81,220) .. (275,220) -- cycle ;
\draw [color={rgb, 255:red, 0; green, 0; blue, 255 }  ,draw opacity=1 ][line width=1.5]    (100,195) .. controls (167.32,134.36) and (168.97,133.27) .. (247.59,193.17) ;
\draw [shift={(250,195)}, rotate = 217.32] [fill={rgb, 255:red, 0; green, 0; blue, 255 }  ,fill opacity=1 ][line width=0.08]  [draw opacity=0] (13.4,-6.43) -- (0,0) -- (13.4,6.44) -- (8.9,0) -- cycle    ;

\draw (107,119) node [anchor=north west][inner sep=0.75pt]  [font=\huge] [align=left] {\textcolor[rgb]{0,0,1}{Direct Effect}};
\draw (107,225.5) node [anchor=north west][inner sep=0.75pt]  [font=\huge] [align=left] {\textcolor[rgb]{1,0,0}{Indirect Effect}};
\draw (61,186.4) node [anchor=north west][inner sep=0.75pt]  [font=\LARGE]  {$X_{1}$};
\draw (162,186.4) node [anchor=north west][inner sep=0.75pt]  [font=\LARGE]  {$X_{2}$};
\draw (262,186.4) node [anchor=north west][inner sep=0.75pt]  [font=\LARGE]  {$X_{3}$};

\end{tikzpicture}